\documentclass[conference]{IEEEtran}
\IEEEoverridecommandlockouts
\usepackage{cite}
\usepackage{amsmath,amssymb,amsfonts}
\usepackage{algorithmic}
\usepackage{graphicx}
\usepackage{textcomp}
\usepackage{xcolor}
\usepackage{subcaption}
\usepackage{times}
\usepackage[justification=centering]{caption}
\usepackage{multicol,lipsum}
\usepackage{multirow}
\usepackage[ruled,vlined]{algorithm2e}
\usepackage[misc]{ifsym}
\usepackage{booktabs}
\def\BibTeX{{\rm B\kern-.05em{\sc i\kern-.025em b}\kern-.08em
    T\kern-.1667em\lower.7ex\hbox{E}\kern-.125emX}}
\begin{document}

\title{\emph{PathRank}: A Multi-Task Learning Framework to Rank Paths in Spatial Networks
}

\author{
\IEEEauthorblockN{Sean Bin Yang and Bin Yang}
\IEEEauthorblockA{\textit{Department of Computer Science, Aalborg University, Denmark} \\
\{seany, byang\}@cs.aau.dk}

}

\maketitle

\begin{abstract}

Modern navigation services often provide multiple paths connecting the same source and destination for users to select. Hence, ranking such paths becomes increasingly important, which directly affects the  service quality. 
We present \emph{PathRank}, a data-driven framework for ranking paths based on historical trajectories using multi-task learning. 
If a trajectory used path $P$ from source $s$ to destination $d$, \emph{PathRank} considers this as an evidence that $P$ is preferred over all other paths from $s$ to $d$. 
Thus, 
a path that is similar to $P$ should have a larger ranking score than a path that is dissimilar to $P$. 
Based on this intuition, \emph{PathRank} models path ranking as a regression problem, where each path is associated with a ranking score. 

To enable \emph{PathRank}, we first propose an effective method to generate a compact set of training data---for each trajectory, we generate a small set of diversified paths.  
Next, we propose a multi-task learning framework to solve the regression problem. In particular, a spatial network embedding is proposed to embed each vertex to a feature vector by considering both road network topology and spatial properties, such as distances and travel times. Since a path is represented by a sequence of vertices, which is now a sequence of feature vectors after embedding, recurrent neural network is applied to model the sequence. The objective function is designed to consider errors on both ranking scores and spatial properties, making the framework a multi-task learning framework. 
Empirical studies on a substantial trajectory data set offer insight into the designed properties of the proposed framework and indicating that it is effective and practical. 
%

%
\end{abstract}



\section{Introduction}

Vehicular transportation reflects the pulse of a city. It 
not only affects people's daily lives and also plays an essential role in many businesses as well as society as a whole~\cite{b1,b2}. With recent deployment of sensing technologies and continued digitization, large amounts of vehicle trajectory data are collected, which provide a solid data foundation to improve the quality of a wide variety of transportation services, such as vehicle routing, traffic prediction, and urban planning. 

A fundamental functionality in vehicular transportation is routing. Given a source and a destination, classic routing algorithms, e.g., Dijkstra's algorithm, identify an optimal path connecting the source and the destination, where the optimal path is often the path with the least travel cost, e.g., the shortest path or the fastest path.  
However, a routing service quality study~\cite{b3} shows that local drivers often choose paths that are neither shortest nor fastest, rendering classic routing algorithms often impractical in many real world routing scenarios. 

To contend with this challenge, a wide variety of advanced routing algorithms, e.g., skyline routing~\cite{b4} and k-shortest path routing~\cite{b5}, are proposed to identify a set of optimal paths, where the optimality is defined based on, e.g., pareto optimality or top-$k$ least costs, which provide drivers with multiple candidate paths. 
In addition, commercial navigation systems, such as Google Maps and TomTom, often follow a similar strategy by suggesting multiple candidate paths to drivers, although the criteria for selecting the candidate paths are often confidential. 

Under this context, ranking the candidate paths is essential for ensuring high routing quality.  
Existing solutions often rely on simple heuristics, e.g., ranking paths w.r.t. their travel times. However, travel times may not always be the most important factor when drivers choose paths, as demonstrated in the routing quality study where drivers often do not choose the fastest paths~\cite{b3}. 
In addition, existing solutions often provide the same ranking to all users but ignore distinct preferences which different drivers may have.  

In this paper, we propose a data-driven ranking framework \emph{PathRank}, which ranks candidate paths by taking into account the paths used by local drivers in their historical trajectories. 
More specifically, \emph{PathRank} models ranking candidate paths as a ``{regression}'' problem---for each candidate path, \emph{PathRank} estimates a ranking score for the candidate path. 

The intuition behind \emph{PathRank} is that if a driver used path $P$ from source $s$ to destination $d$, this means that the driver considered path $P$ as the ``best'' path over all possible paths from $s$ to $d$. Then, a path that is similar to $P$ should rank higher than a path that is dissimilar to $P$.  
%
%

Based on the above intuition, 
for each historical trajectory, we identify the path $P$ used by the trajectory and the source $s$ and the destination $d$ of the trajectory. We consider path $P$ as the ground truth path. Next, we identify a set of paths $\mathcal{PS}$ that connect $s$ and $d$. %
For each candidate path $P^\prime\in \mathcal{PS}$, we associate a similarity score $\mathit{sim}(P, P^\prime)$ that measures how similar between the path $P^\prime$ and the ground truth path $P$. Here, a number of path similarity functions~\cite{b6} can be applied as function $\mathit{sim}(\cdot, \cdot)$, e.g., weighted Jaccard similarity~\cite{b7}. 
%

In the training phase, set $\{(P^\prime, sim(P, P^\prime)\}$ is used to train a regression model, where path $P^\prime$ is a training instance and $sim(P, P^\prime)$ is its label, i.e., the ranking score. 
After training, we obtain a regression model. Then, in the testing phase, given a set of candidate paths returned by advanced routing algorithms or Google Maps, the regression model is able to estimate a ranking score for each candidate path. 
Finally, we rank the candidate paths w.r.t. their ranking scores. 

We may train \emph{PathRank} on historical trajectories from a specific driver and thus provide a personalized ranking for the driver. 
Alternatively, we can also train \emph{PathRank} on historical trajectories from many drivers and thus provide a generic ranking, which, for example, can be used for different drivers, especially for new drivers who do not have many or even no historical trajectories. We include an empirical study on this in Section~\ref{sec:exp-driver-specific}.
%

Enabling \emph{PathRank} is non-trivial as we need to face two major challenges. First, constructing an appropriate training path set $\mathcal{PS}$ is non-trivial. Since there may exist a large amount of paths from a source to a destination, it is thus prohibitive to include all such paths in $\mathcal{PS}$. Selecting a small subset of such paths may adversely affect the training effectiveness. 
Thus, it is challenging to select a small, representative subset of paths to be included in $\mathcal{PS}$ such that they enable both \emph{efficient} and \emph{effective} training and thus providing accurate ranking while maintaining efficiency.  

Second, effective regression models often rely on meaningful feature representations of input data. In our setting, the input is a path and no existing methods are available to represent paths in a meaningful feature space to enable ranking. Here, the meaningful feature space should take into account both the topology of the underlying road network and the spatial properties, such as distances and travel times, of the road network.

To contend with the first challenge, we propose an effective method to generate a compact training path set $\mathcal{PS}$. We consider different travel costs that drivers may consider, e.g., distance, travel time, and fuel consumption. Next, for each travel cost, we identify a set of \emph{diversified}, top-$k$ least-cost paths. Here, two paths are diversified if the path similarity between them is smaller than a threshold, e.g., 0.8, where a number of different path similarity functions can be applied here as well~\cite{b6}. 
As an example, diversified top-3 shortest paths consist of three paths where the path similarity of every pair of paths is smaller than a threshold and there does not exist another set of three paths which are mutually diversified and whose total distance is shorter. 
Considering diversity avoids including top-3 shortest paths where they only differ slightly, e.g., one or two edges. 
This method makes sure that the candidate path set (i) considers multiple travel costs that a driver may consider when making routing decisions; and (ii) includes paths that are dissimilar with each other, which in turn represent a large feature space of the underlying road network. 

Next, we propose a deep learning framework to learn meaningful feature representations of paths which enables effective ranking and thus solve the second challenge. 
Recall that the input is a path, which is represented as a sequence of vertices in a road network graph. 
To capture the graph topology, we utilize unsupervised graph embedding, e.g.,  node2vec~\cite{b8}, to transform a vertex into a feature vector. 
Since a path is a sequence of vertices, we employ a recurrent neural network (RNN) to model the sequence of the corresponding feature vectors of the vertices. 
So far, thanks to the graph embedding, the framework considers the topology of the underlying road network, but we also need to consider spatial properties of the road network, which are not captured by classic graph embedding. To this end, we let the RNN not only estimate the similarity w.r.t. the ground truth path but also reconstruct the path's spatial properties, such as the length, the travel time, and the fuel consumption of the path. This makes the framework a \emph{multi-task} learning framework where the \emph{main task} is to estimate the similarity which is used for the final ranking and the \emph{auxiliary tasks} are to enforce the graph embedding to also capture the spatial properties of the underlying road network which eventually improve the accuracy of the main task.  

To the best of our knowledge, this is the first data-driven, end-to-end solution for ranking paths in spatial networks. Specifically, we make four contributions. First, we propose a method to generate a compact set of training paths which enables effective and efficient learning. 
Second, we propose a multi-task learning framework to enable spatial network embedding that enhances classic graph embedding by incorporating spatial properties.  
Third, we integrate the spatial network embedding with similarity regression to provide an end-to-end solution for ranking paths. 
Fourth, we conduct extensive experiments using a large real world trajectory set to offer insight into the design properties of the proposed framework and to demonstrate that the framework is effective. 

Paper Outline: Section 2 covers related work. Section 3 covers preliminaries. Section 4 discusses how to generate the training data. Section 5 proposes \emph{PathRank}, including basic framework and advanced framework. Section 6 reports on empirical evaluations. Section 7 concludes.
\section{Related Work}
We review related studies on learning to rank in the context of information retrieval, graph representation learning, and trajectory learning.  

\begin{figure}
\centerline{\includegraphics[scale=0.5]{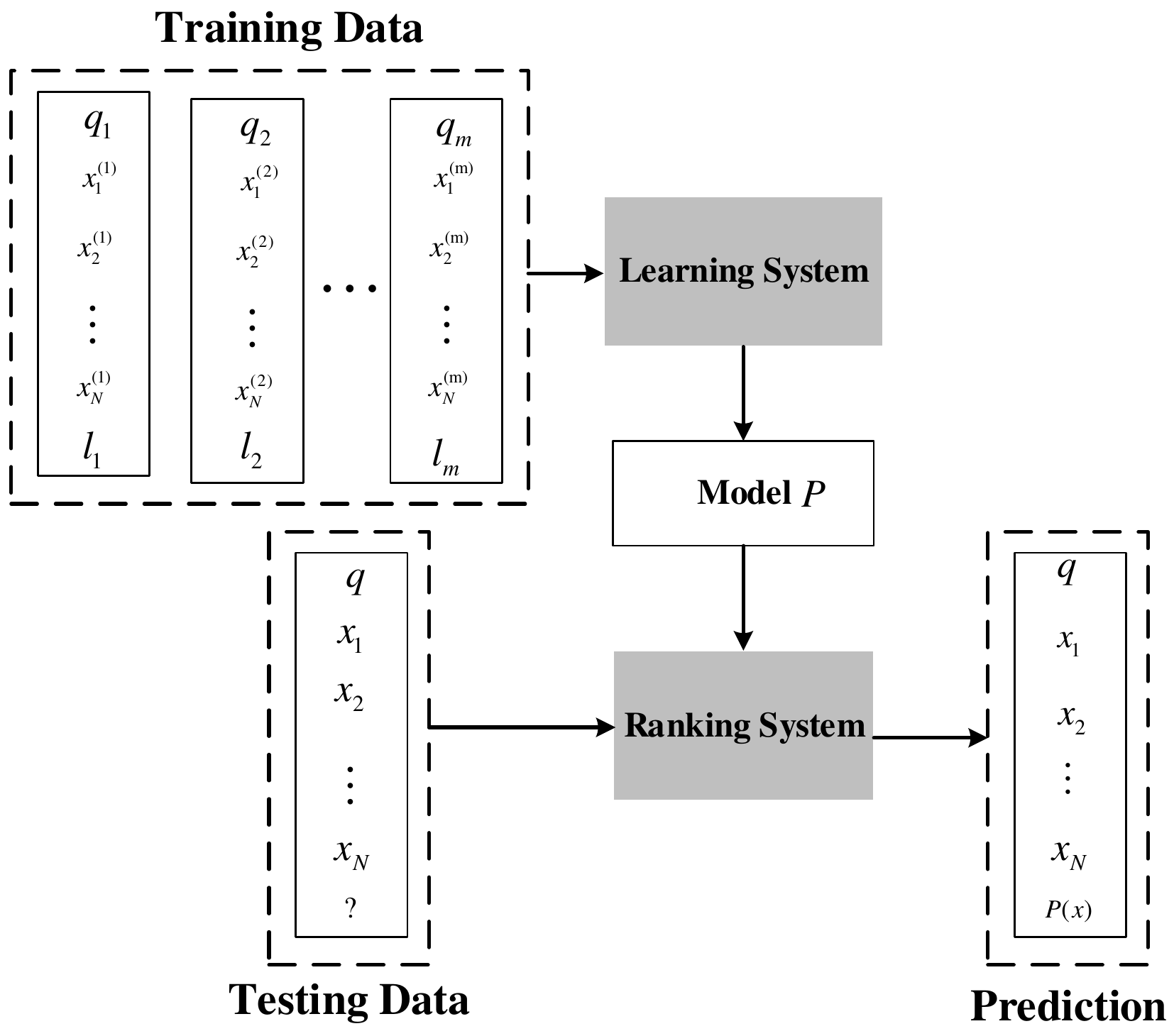}}
  \caption{Learn to rank Framework.}
  \label{fig:LTR}
\end{figure}

\noindent
\textbf{Learning to rank}
%
Learning to rank plays an important role in ranking in the context of information retrieval (IR), where the primary goal is to learn how to rank documents or web pages w.r.t. queries, which are all represented as feature vectors. Fig.~\ref{fig:LTR} gives the typical learning to rank framework.   
%
%
Learning to rank methods in IR can be categorized into point-wise, pair-wise, and list-wise methods. 
%
Point-wise methods estimate a ranking score for each individual document. Then, the documents can be ranked based on the ranking scores~\cite{b9}. 
Pair-wise methods focus on, for a given pair of documents, making a binary decision on which document is better, i.e., a relative order. Here, although we do not know the ranking scores for individual documents, we are still able to rank documents based on the estimated relative orders~\cite{b10, b11}. 
List-wise methods take into account a set of documents and estimate the ranking for the documents~\cite{b12}. 
Recently, deep learning is applied in learning to ranking in IR with a focus on learning semantic meaningful representation of both queries and documents, such as DSSM~\cite{b13}, CDSSM~\cite{b14} and DeepRank~\cite{b15}.

Although learning to rank techniques have been applied widely and successfully in IR, they only consider textual documents and queries and cannot be applied for ranking paths in spatial networks, since both graph topology and spatial properties, which are the two most important factors in spatial networks, are ignored. 
We follow the idea of the point-wise learning to rank techniques in IR and propose \emph{PathRank} to rank paths in spatial networks while considering both graph topology and spatial properties. 

\noindent
\textbf{Network Representation Learning}
%
Network representation learning, a.k.a., graph embedding, aims to learn low-dimensional feature vectors for vertices while 
preserving network topology structure 
such that the vertices with similar feature vectors share similar structural properties~\cite{b8,b16,b17,b18,b20,b21}. 
We distinguish two categories of methods: random walk based methods and deep learning based methods. 

A representative method in the first category is DeepWalk~\cite{b17}. DeepWalk first samples sequences of vertices based on truncated random walks, where the sampled vertex sequences capture the connections between vertices in the graph. 
Then, skip-gram model~\cite{b19} is used to learn low-dimensional feature vectors based on the sampled vertex sequences. 
Node2vec~\cite{b8} considers higher order proximity between vertices by maximizing the probability of occurrences of subsequent vertices in fixed length random walks. 
A key difference from DeepWalk is that node2vec employs biased-random walks that provide a trade-off between breadth-first  and depth-first searches, and hence achieves higher quality and more informative embedding than DeepWalk does. 

To overcome the weaknesses of random walk based methods, e.g., the difficulty in determining the random walk length and the number of random walks, deep learning based methods utilize the random surfing model to capture contextual relatedness between each pair of vertices and preserves them into low-dimensional feature vectors for vertices~\cite{b20}. 
Deep learning based methods are also able to 
take into account complex non-linear relations. 
GraphGAN~\cite{b21} is proposed to learn vertex representations by modeling the connectivity behavior through an adversarial learning framework using a minimax game.

LINE~\cite{b18} does not fall into the above two categories. 
Instead of exploiting random walks to capture network structures, LINE~\cite{b18} propose a model with a carefully designed objective function that preserves both the first-order and second-order proximities.   

However, all existing graph embedding methods consider non-spatial networks such as social networks, citation networks, and biology networks. They ignore spatial properties, e.g., distances and travel times, which are crucial features in spatial networks such as road networks. 
%
%
%
In this paper, we propose a multi-task learning framework to extend existing graph embedding to incorporate important spatial properties. Experimental results show that the graph embedding that considers spatial-properties gives the best performance when ranking paths in spatial networks.  

\noindent
\textbf{Trajectory Learning }
Machine learning have been also applied on trajectories to support different applications~\cite{DBLP:journals/pvldb/0002GJ13,DBLP:conf/icde/HuG0J19,DBLP:journals/tkde/YangKJ14}. A multi-task learning framework~\cite{DBLP:conf/cikm/Kieu0GJ18} is proposed to distinguish trajectories from different drivers. When considering trajectories as time series, recurrent autoencoders~\cite{DBLP:conf/mdm/Kieu0J18,DBLP:conf/cikm/CirsteaMMG018} and recurrent autoencoder ensembles~\cite{IJCAI}  are proposed to identify outliers. However, these studies do not take into account underlying road network structures into consideration. 

In addition, different approaches have been proposed to learn personalized driving preferences from trajectories~\cite{DBLP:conf/icde/DaiYGD15,DBLP:conf/icde/Guo0HJ18,DBLP:journals/vldb/YangDGJH18}, which enable personalized routing. However, in this paper, we consider an orthogonal approach where we rank candidate paths, which can be obtained from well-known navigation services, rather than proposing yet another personalized routing approach which may not be easily integrated with existing navigation services. 

Finally, trajectories have been applied to extract high-resolution travel costs~\cite{b4,DBLP:journals/geoinformatica/HuYJM17,DBLP:journals/vldb/HuYGJ18,DBLP:journals/vldb/YangDGJH18,DBLP:journals/pvldb/DaiYGJH16}, such as travel time and fuel consumption~\cite{DBLP:journals/geoinformatica/Guo0AJT15,DBLP:conf/gis/GuoM0JK12}. In particular, time-varying and uncertain travel costs can be learned from trajectories. It is of interest to extend \emph{PathRank} to consider time-varying and uncertain traffic conditions as future work.

\section{Preliminaries}

\subsection{Basic Concepts}
A \emph{road network} is modeled as a weighted, directed graph $G = (\mathbb{V}, \mathbb{E}, D, T, F)$. Vertex set $\mathbb {V}$ represents road intersections and road ends; edge set ${\mathbb {E}} \subset \mathbb{V} \times \mathbb{V}$ represents road segments. 
Functions $D$, $T$, and $F$ maintain the travel costs of the edges in graph $G$. Specifically, function $D: \mathbb{E}\rightarrow \mathbb{R}^+$ maps each edge to its length. Functions $T$ and $F$ have similar signatures and maps edges to their travel times and fuel consumption, respectively. 



A \emph{path} ${\boldsymbol {P} = \left(v_{1},v_{2},v_{3},\ldots,v_{X}\right)}$ is a sequence of $X$ vertices where $X > 1$ and each two adjacent vertices must be connected by an edge in $\mathbb{E}$.  


A \emph{trajectory} ${\boldsymbol {T} = \left(p_1,p_2,p_3,\ldots,p_Y\right)}$ is a sequence of GPS records pertaining to a trip, where each GPS record $p_i=(location, time)$ represents the location of a vehicle at a particular timestamp. The GPS records are ordered according to their corresponding timestamps, where $p_i.time <p_j.time$ if $1 \leq i<j \leq Y$. 

Map matching~\cite{b22} is able to map a GPS record to a specific location on an edge in the underlying road network, thus aligning a trajectory with a path in the underlying road network. We call such paths \emph{trajectory paths}. 
In addition, a trajectory $\boldsymbol {T}$ is also associated with a driver identifier, denoted as  $\boldsymbol {T.driver}$, indicating who made the trajectory. 





Multiple similarity functions~\cite{b1,b6,b7,b23} are available to calculate the similarity between two paths, where the most popular functions belong to the Jaccard similarity function family, in particular, the weighted Jaccard similarity~\cite{b1,b7}. 
%
In this paper, we use the weighted Jaccard Similarity (see Equation~\ref{eq:wj}) to evaluate the similarity between two paths. However, other similarity functions can be easily incorporated into the proposed framework. 

\begin{align}
\label{eq:wj}
 \operatorname{{\textit{sim}}}\left(P_{1}, P_{2}\right)=\frac{\sum_{e \in P_{1} \cap P_{2}} {G.D}(e)}{\sum_{e \in P_{1} \cup P_{2}} {G.D}(e)}
\end{align}
%
%
Here, we use $P_{1} \cap P_{2}$ and $P_{1} \cup P_{2}$ to represent two edge sets: edge set $P_{1} \cap P_{2}$ consists of the edges that appear in both $P_1$ and $P_2$; and edge set $P_{1} \cup P_{2}$ consists of the edges that appear in either $P_1$ or $P_2$. Recall that function ${G.D}(e)$ returns the length of edge $e$. 
Then, the intuition of the weighted Jaccard similarity is two-fold: first, the more edges the two paths share, the more similar the two paths are; second, the longer the shared edges are, the more similar the
two paths are.

\subsection{PathRank Overview}

\begin{figure}
\centerline{\includegraphics[scale=0.6]{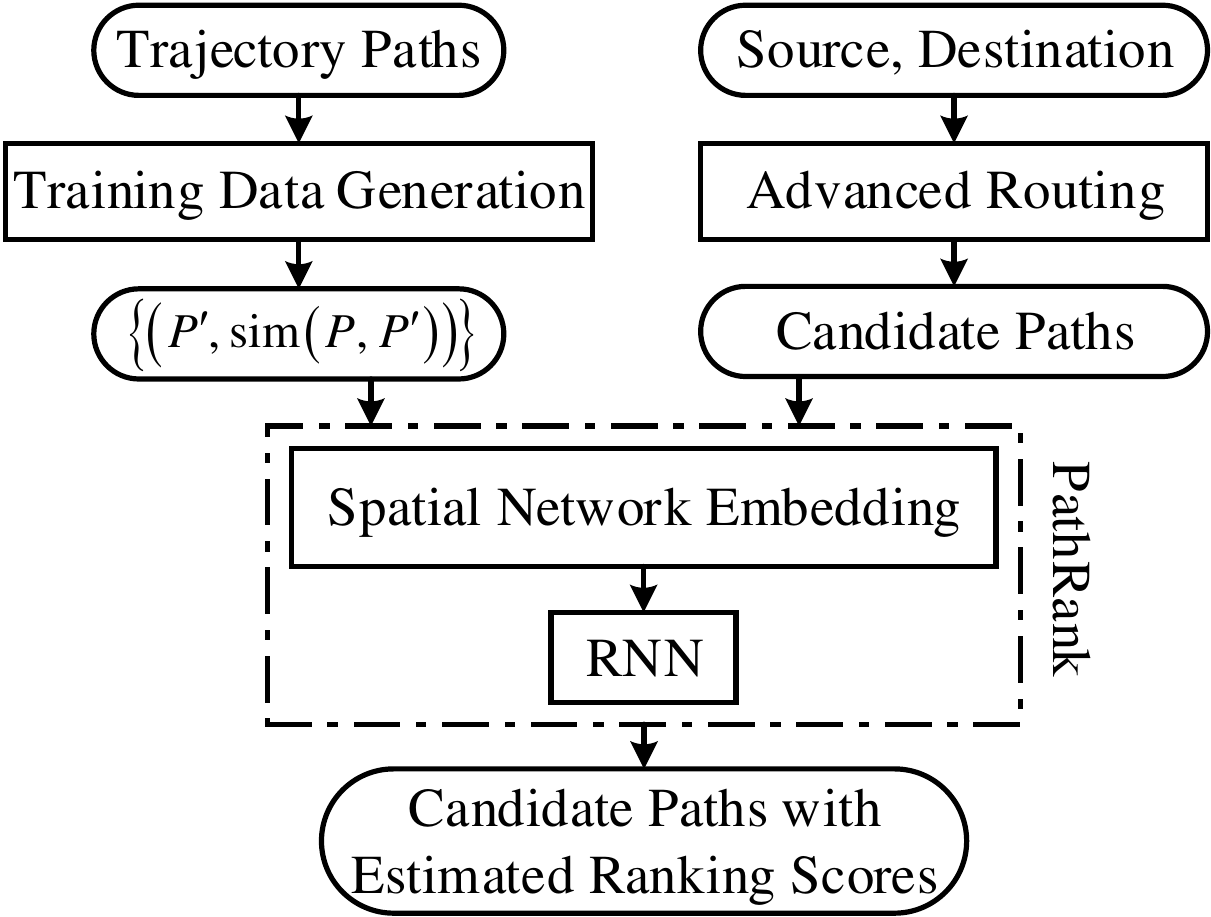}}
  \caption{Solution Overview.}
  \label{fig:overview}
  \vspace{-10pt}
\end{figure}

Fig.~\ref{fig:overview} shows an overview of the proposed \emph{PathRank}. 
Given a set of historical trajectory, we first map match them to obtain their corresponding \emph{trajectory paths}. 
In the training phase, the trajectory paths are fed into the \emph{Training Data Generation} module. For each trajectory path $P$,  
the training data generation module generates a compact set $\mathcal{PS}$ of competitive paths such that each competitive path $P^\prime\in \mathcal{PS}$ also connects the same source and destination of the trajectory path $P$. 
Next, we consider the trajectory path $P$ as the ground truth path and thus compute a similarity score $\mathit{sim}(P, P^\prime)$ for each competitive path $P^\prime$. 
The training data generation module iterates over each trajectory path $P$ and generates competitive paths along with similarity scores. The output of the module is a set of ``competitive path'' and ``similarity score'' pairs, denoted as $\{(P^\prime, sim(P, P^\prime)\}$, which is used as the input for the \emph{PathRank}.  

In the training phase, for each training instance $(P^\prime, sim(P, P^\prime))$, the \emph{Spatial Network Embedding Module} embeds each vertex in competitive path $P^\prime$ into a feature vector. This transfers path $P^\prime$ into a sequence of feature vectors, which is then fed into a \emph{Recurrent Neural Network} (RNN). 
The RNN estimates the similarity between ground truth trajectory path $P$ and path $P^\prime$. An objective function is designed to measure the discrepancy between the estimated similarity and the ground truth similarity $sim(P, P^\prime)$. Then, the whole training process aims to minimize the objective function. 

In the testing phase, we use the trained \emph{PathRank} to rank candidate paths. Given a source and a destination, advanced routing algorithms or commercial navigation systems are able to provide multiple candidate paths, which are used as testing instances.  
Next, \emph{PathRank} takes as input each testing path and returns an estimated ranking score. Finally, we are able to rank the testing paths according to their estimated ranking scores.

\section{Training Data Generation}


We proceed to elaborate how to generate a compact set of training paths for a trajectory path. 


\subsection{Intuitions}

Ranking paths is similar to rank products in online shops. 
If a user clicks a specific product, it provides evidence that the user is interested in the product than other similar products. 
Similarly, a trajectory path $P$ from a source $s$ to destination $d$ also provides evidence that a driver prefers path $P$ than other paths that connect $s$ to $d$. 

The main difference is that, in online shops, the other similar products, i.e., competitor products, can be obtained explicitly, e.g., those products that are shown to the user in the same web page but are not clicked by the user.  
Based on the positive and negative training data, i.e., the products that are clicked and not clicked by the user, effective learning mechanism, e.g., learning to rank~\cite{b9, b10, b11, b12, b13, b14,b15}, is available to learn an appropriate ranking function.  

However, in our setting, the other candidate paths are often unknown and implicit because we do not know when the driver made the decision to take path $P$, what other paths were in driver's mind. 
Thus, the main target of the training data generation module is to generate a set of paths $\mathcal{PS}$ which may include the other paths when the driver made decision to use trajectory path $P$. We call $\mathcal{PS}$ \emph{competitive path set}. 

A naive way to generate the competitive path set is to simply include all paths from $s$ to $d$. This is infeasible to use in real world settings since the competitive path set may contain a huge number of paths in a city-level road network graph, which in turn makes the training prohibitively inefficient. Thus, we aim to identify a \emph{compact} competitive path set, where only a small number of paths, e.g., less than 10 paths, are included. 




\begin{figure*}
     \centering
     \begin{subfigure}[b]{0.32\textwidth}
         \centering
         \includegraphics[width=\textwidth,height=3.6cm]{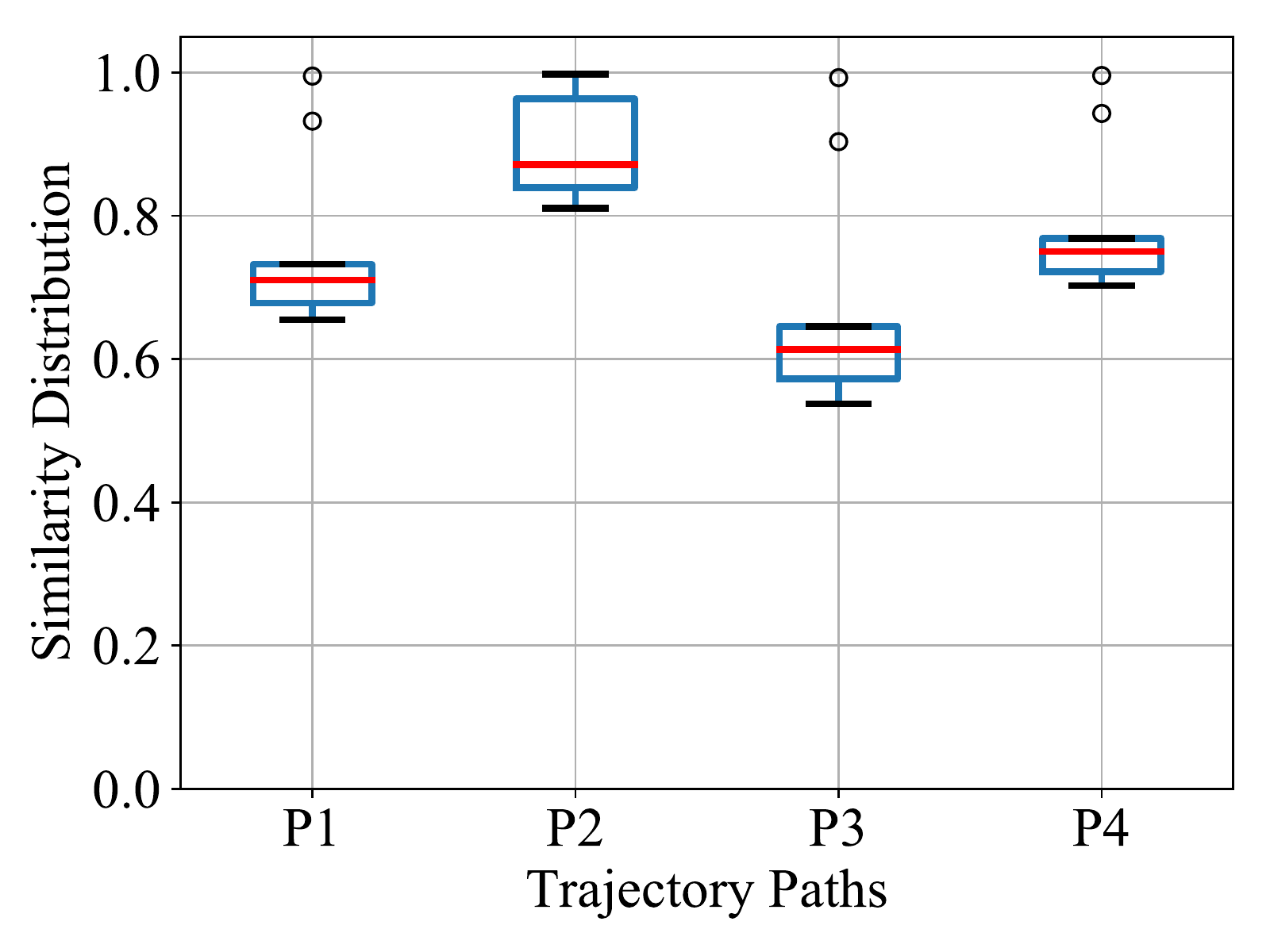}
         \caption{\emph{TkDI}}
         \label{fig:subfig:tkdi}
     \end{subfigure}
     \hfill
     \begin{subfigure}[b]{0.32\textwidth}
         \centering
         \includegraphics[width=\textwidth,height=3.6cm]{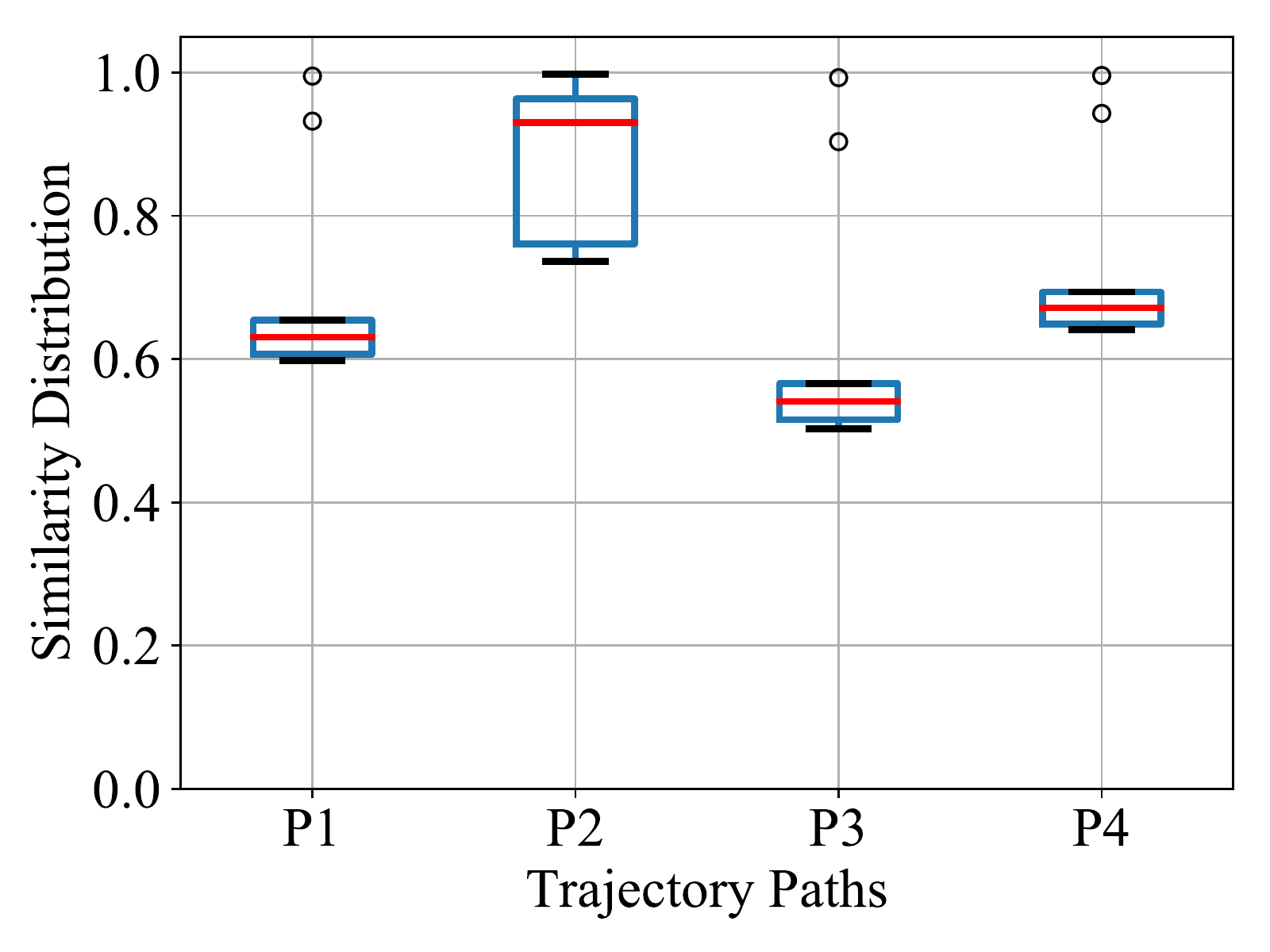}
         \caption{\emph{TkTT}}
         \label{fig:subfig:tktt}
     \end{subfigure}
     \hfill
     \begin{subfigure}[b]{0.32\textwidth}
         \centering
         \includegraphics[width=\textwidth,height=3.6cm]{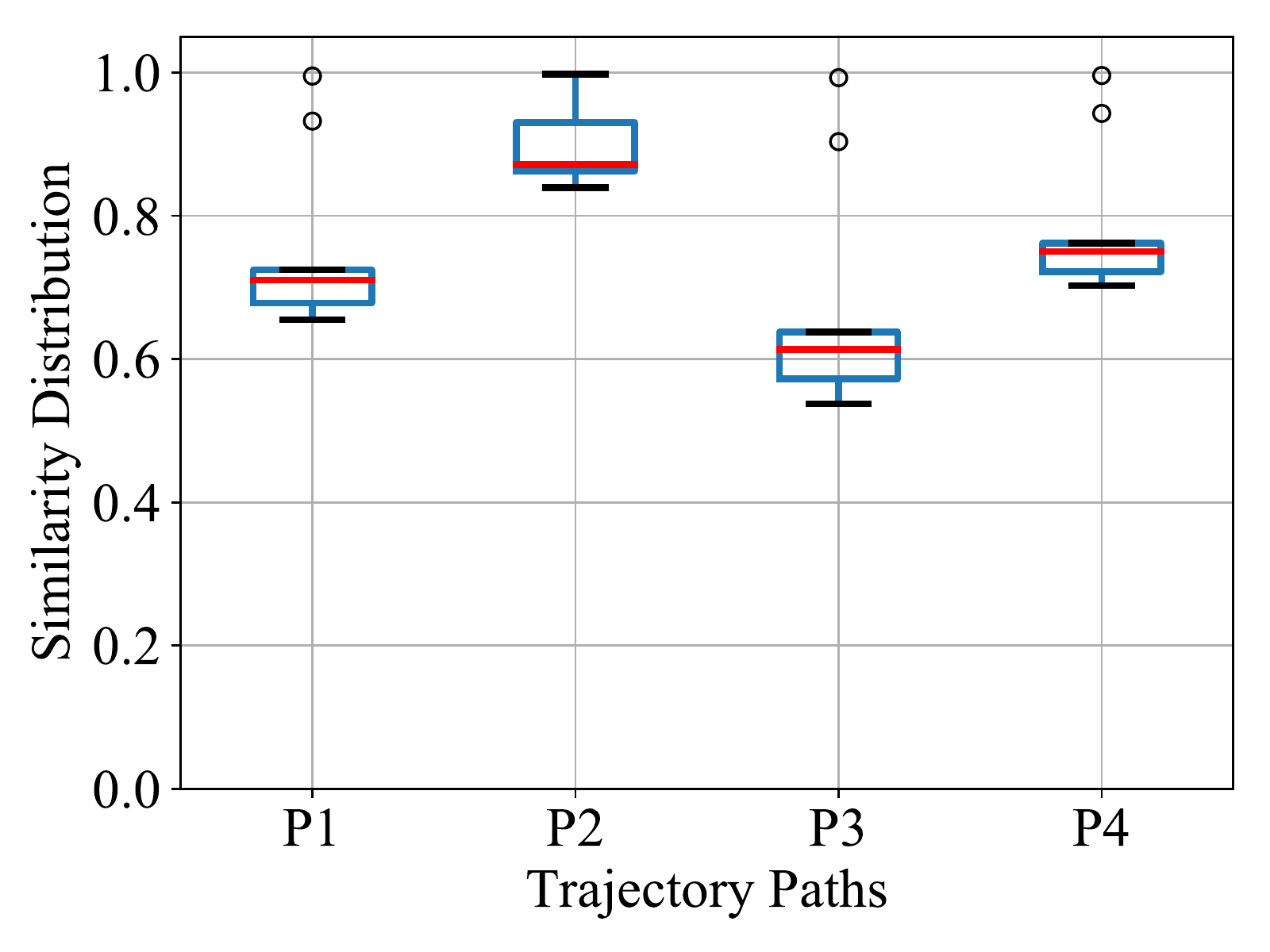}
         \caption{\emph{TkFC}}
         \label{fig:subfig:tkfc}
     \end{subfigure}
     \hfill
     \begin{subfigure}[b]{0.32\textwidth}
         \centering
         \includegraphics[width=\textwidth,height=3.6cm]{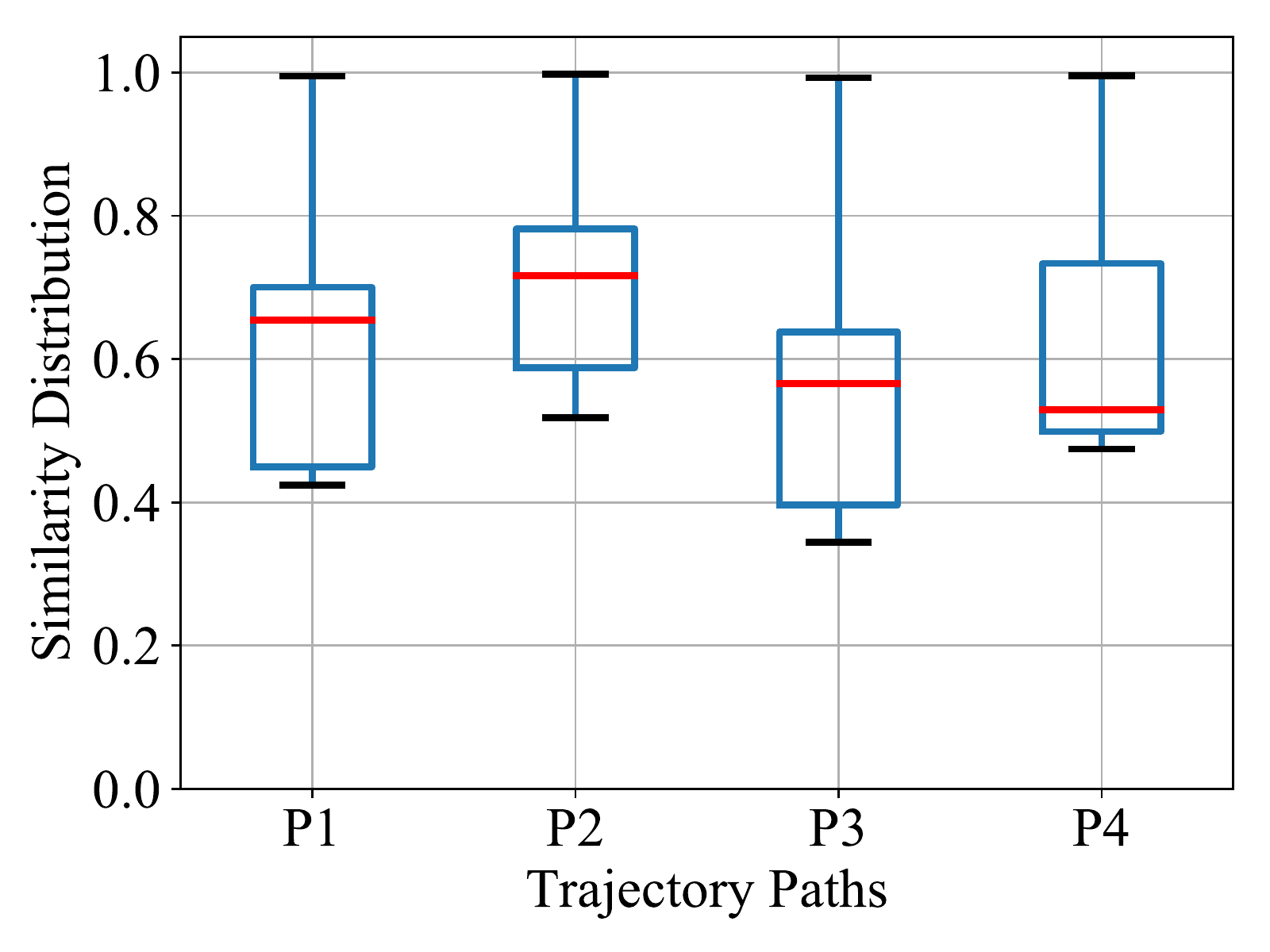}
         \caption{\emph{D-TkDI}}
         \label{fig:subfig:dtkdi}
     \end{subfigure}
     \hfill
     \begin{subfigure}[b]{0.32\textwidth}
         \centering
         \includegraphics[width=\textwidth,height=3.6cm]{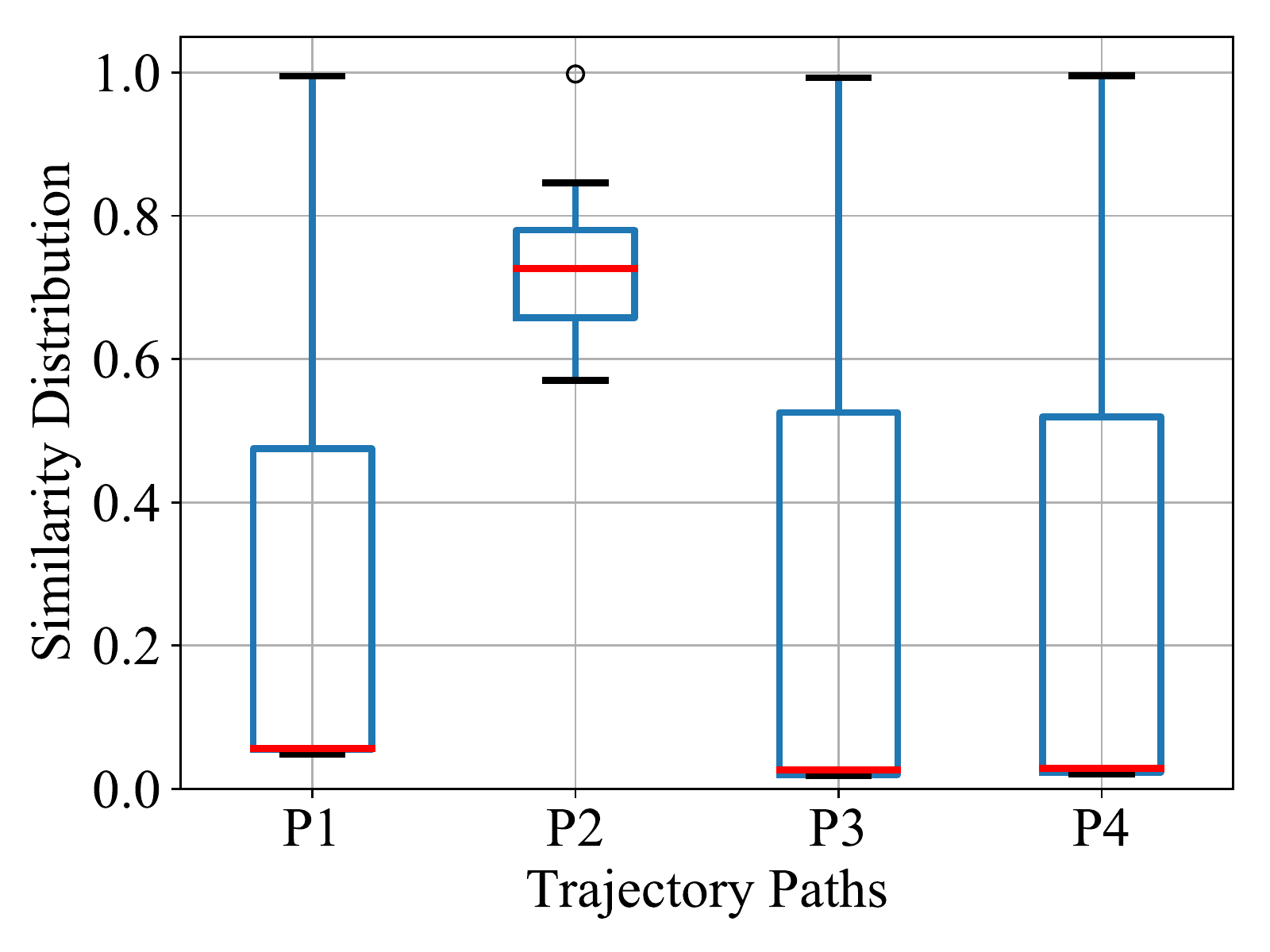}
         \caption{\emph{D-TkTT}}
         \label{fig:subfig:dtktt}
     \end{subfigure}
     \hfill
     \begin{subfigure}[b]{0.32\textwidth}
         \centering
         \includegraphics[width=\textwidth,height=3.6cm]{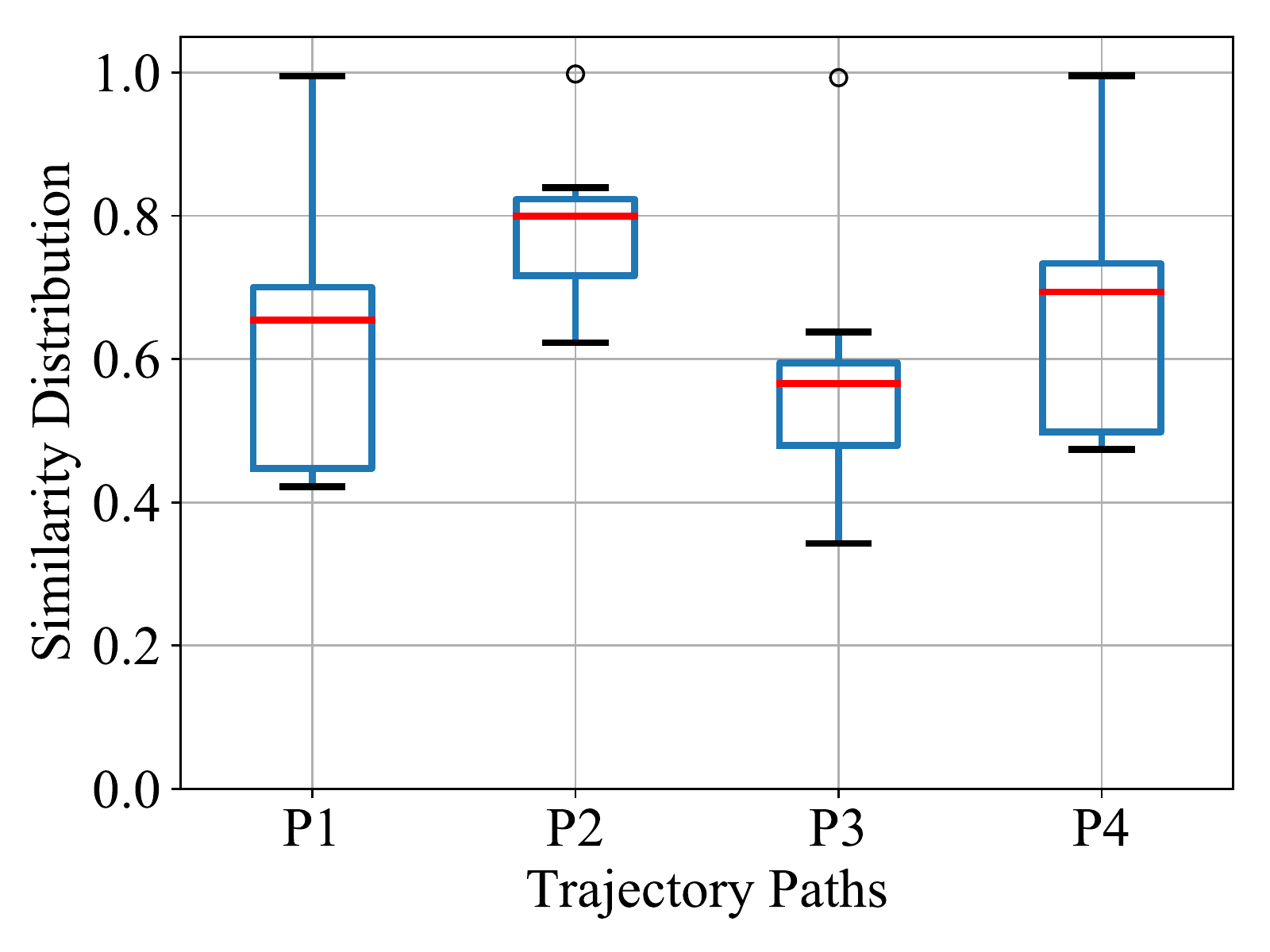}
         \caption{\emph{D-TkFC}}
         \label{fig:subfig:dtkfc}
     \end{subfigure}
        \caption{Top-$k$ Paths vs. Diversified Top-$k$ Paths, k=9.}
        \label{fig: boxplot}
\end{figure*}

\subsection{Top-$k$ Shortest Paths}

The first strategy is to employ a classic top-$k$ shortest path algorithm, e.g., Yen's algorithm~\cite{b24}, to include the top-$k$ shortest paths from $s$ to $d$ into the competitive path set $\mathcal{PS}$. 



This strategy is simple and efficient since a wide variety of efficient algorithms are available to generate top-$k$ shortest paths in the literature~\cite{b24, b25, b26, b27}. 
However, a serious issue of this strategy is that the top-$k$ shortest paths are often highly similar. Thus, their similarities w.r.t. the ground truth, trajectory path $P$, are also similar, which adversely affect the effectiveness of the subsequent ranking score regression. 


For example, we choose four trajectory paths with different sources and destinations. For each trajectory path, we generate top-9 shortest, fastest, and most fuel-efficient paths connecting the same source and destination as the competitive paths. Then, we compute the competitive paths' similarities w.r.t. the trajectory path. Figures~\ref{fig:subfig:tkdi}, ~\ref{fig:subfig:tktt}, and ~\ref{fig:subfig:tkfc} show the box plots of the similarities per trajectory path. We observe that the similarities often only spread over a very small range. For example, for the first trajectory path $P_1$, its corresponding top-9 shortest paths have similarities spreading from 0.65 to 0.75.   

If the similarities of competitive paths only spread over a small range, they only provide training instances for estimating ranking scores in the small range, which may make the trained model unable to make accurate estimations for ranking scores outside the small range. Thus, an ideal strategy should be providing a set of competitive paths whose similarities cover a large range. To this end, we propose the second strategy using the diversified top-$k$ shortest paths.

%
\subsection{Diversified Top-$k$ Shortest Paths}

Diversified top-$k$ shortest paths finding aims at identifying top-$k$ shortest paths such that the paths are mutually dissimilar, or diverse, with each other. 
%

Algorithm~\ref{alg:DkPS} details the procedure of finding diversified top-$k$ shortest path. 
First, we always include the shortest path into the diversified top-$k$ shortest path set $\mathit{DkPS}$. 
Next, we get into a loop where we keep checking the next shortest path $P_i$ until we have included $k$ paths in $\mathit{DkPS}$ or we have checked all paths connecting the source and destination. 
When checking the next shortest path $P_i$, we include $P_i$ into $\mathit{DkPS}$ if the similarity between $P_i$ and each existing path in $\mathit{DkPS}$ is smaller than a threshold $\delta$. This means that $P_i$ is sufficiently dissimilar with the paths in $\mathit{DkPS}$, thus making sure that $\mathit{DkPS}$ is a  diverse top-$k$ shortest path set. 
The smaller the threshold $\delta$ is, the more diverse the paths in $\mathit{DkPS}$ are. However, if the threshold $\delta$ is too small, it may happen that less than $k$ diverse shortest paths or even only the shortest path are included in $\mathit{DkPS}$. 



 
\begin{algorithm}
\caption{Top-$k$ Diversified Paths}
\label{alg:DkPS}
\LinesNumbered
\KwIn{Road network $G$, source $s$, destination $d$, integer $k$, similarity threshold $\delta$} 
\KwOut{The diversified top-$k$ paths: $\mathit{DkPS}$}
Add the shortest path $P_1$ into $\mathit{DkPS}$\;
\While{$\mathit{DkPS}<$ k} {
 Identify the next shortest path $P_{i}$ \;
 Boolean $\mathit{tag} \leftarrow$ true; \\
  \For{each path $P \in \mathit{DkPS}$}
  {
    \If{$\operatorname{sim}\left(P_{i}, P\right)$ $\geq$ $\delta$}
    {
        $\mathit{tag} \leftarrow$ false; \\
        \textbf{Break};
    }
  }
  \If{tag}
  {
    Add $P_i$ into $\mathit{DkPS}$\;
  }
}
\textbf{return} $\mathit{DkPS}$\;
\end{algorithm}

Figures~\ref{fig:subfig:dtkdi}, ~\ref{fig:subfig:dtktt}, and ~\ref{fig:subfig:dtkfc} show the box plots of the similarities of the same four trajectory paths when using diversified top-$9$ shortest, fastest, and most fuel efficient paths with threshold $\delta=0.8$. We observe that the similarities spread over larger ranges compared to Figure~\ref{fig:subfig:tkdi}, ~\ref{fig:subfig:tktt}, and ~\ref{fig:subfig:tkfc} when using classic top-$k$ shortest paths. 

\subsection{Considering Multiple Travel Costs}

Recent studies on personalized routing~\cite{b1, b7} suggest that a driver may consider different travel costs, e.g., travel time, distance, and fuel consumption, when making routing decisions. 
This motivates us to consider multiple travel costs, but not only distance, when generating competitive path sets. 
The first option to do so is to use Skyline routing~\cite{b4}, which is able to identify a set of pareto-optimal paths, a.k.a., Skyline paths, when considering multiple travel costs.  However, Skyline routing also suffers the high similarity problem that the classic top-$k$ shortest paths have---it often happens that the skyline paths are mutually similar, which may adversely affect the training effectiveness. 

We propose a simple yet effective approach. We run the diversified top-$k$ shortest paths $x$ times where each time we consider a specific travel cost. Then, we use the union of the diverse paths as the final competitive path set $\mathit{PS}$. 
For example, when considering three travel costs, i.e., distances, travel times, and fuel consumption, we set $x=3$ and identify the diversified top-$k$ shortest, fastest, and most fuel efficient paths, respectively. Then, the union of the diversified top-$k$ shortest, fastest, and most fuel efficient paths is used as the final competitive path set $\mathit{PS}$.

Since we run the diversified top-$k$ shortest path finding multiple times for different travel costs, we can use a small $k$ for each run. For example, when we set $k=3$ and consider three travel costs, this makes $\mathit{PS}$ also consist of up to 9 paths including the top-3 shortest, fastest, and most fuel efficient paths. 

To summarize, we use multi-cost, diversified top-$k$ least-cost paths as the compact competitive path set $\mathit{PS}$ for each trajectory path. Next, we combine the competitive path sets from all trajectory paths together to obtain a set of ``competitive path'' and ``similarity score'' pairs, denoted as $\{(P^\prime_i, \mathit{sim}_i)\}$. Here, competitive path $P^\prime_i$ is the input instance and similarity score $\mathit{sim}_i$ is the corresponding label. This set is used as the training data for \emph{PathRank}.


\section{PathRank}

We propose an end-to-end deep learning framework to estimate similarity scores for paths. We first propose a basic framework that consists of a vertex embedding network and a recurrent neural network. Next, we extend the vertex embedding network to capture both the topology and spatial properties of a road network graph, which improves the learning accuracy. 


\subsection{Basic Framework}

\begin{figure*}[ht]
\includegraphics[width=\textwidth]{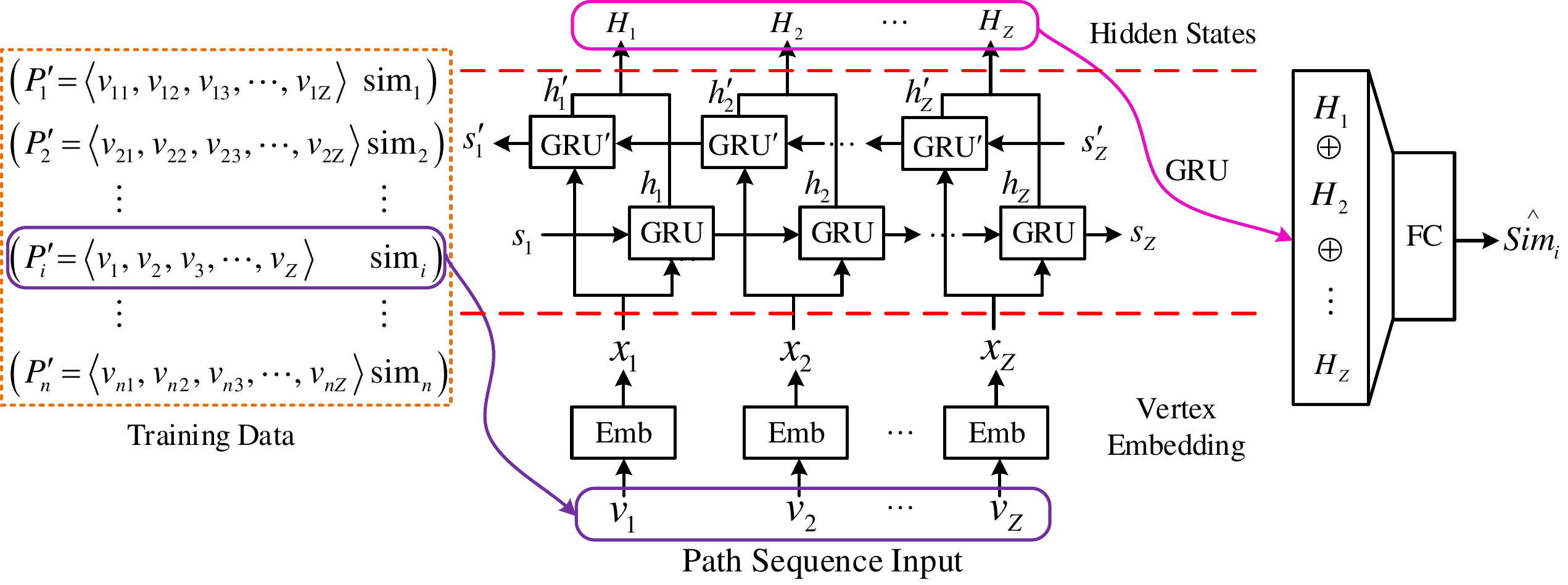}
  \caption{Basic Framework of \emph{PathRank}.}
  \label{fig:basicPR}
\end{figure*}

%
Recall that the input for \emph{PathRank} is a path, i.e., competitive path $P^\prime_i$, and the label of the input is its similarity score $\mathit{sim}_i$.  
In order to use deep learning to solve the similarity score regression problem, a prerequisite is to represent the input path $P^\prime_i$ into an appropriate feature space. 
To this end, we propose to use a vertex embedding network to transfer each vertex in the input path to a feature vector. Since a path is a sequence of vertices, after vertex embedding, the path becomes a sequence of feature vectors. 
RNN finally captures the features of path sequence, which is applied to compute an estimated similarity score.
Next, since RNNs are capable of capturing dependency for sequential data, we employ an RNN to model the sequence of feature vectors. The RNN finally outputs an estimated similarity score, which is compared against the ground truth similarity $\mathit{sim}_i$. 
This results in the basic framework of \emph{PathRank}, which consists of two neural networks---a vertex embedding network and a recurrent neural network (RNN), as shown in Figure~\ref{fig:basicPR}. 

\subsubsection{Vertex Embedding}
We represent a vertex $v_i$ in road network graph $G$ as a one-hot vector $q_i\in \mathbb{R}^{N}$, where $N$ represents the number of vertices in $G$, i.e., $N=|G.\mathbb{V}|$. 
Specifically, the $i$-th vertex $v_i$ in graph $G$ is represented as a vector $q_i$ where the $i$-th bit is 1 and the other $N-1$ bits are 0.  

Vertex embedding employs an embedding matrix $B \in \mathbb{R}^{M \times N}$ to transfer a vertex's one-hot vector $q_i$ into a new feature vector $x_i=B q_i\in \mathbb{R}^{M}$. 
The feature vector is often in a smaller space, where $M<N$. 

Given a competitive path $P^\prime_i=\langle v_{1}, v_{2}, \ldots, v_{Z}\rangle$, we apply the same embedding matrix $B$ to transfer each vertex to a feature vector. Thus, the competitive path $P$ is represented as a sequence of features $\langle x_{1}, x_{2}, \ldots, x_{Z}\rangle$, where $x_j=B  q_j$ and $1\leq j\leq Z$.   


\subsubsection{RNN}

The feature sequence represents the flow of travel on path $P^\prime_i$ and we would like to capture the flow. To this end, we fed the feature sequence $\langle x_{1}, x_{2}, \ldots, x_{Z}\rangle$ into a recurrent neural network, which is known to be effective for modeling sequences. 
Specifically, we employ a bidirectional gated recurrent neural network (BD-GRU) to capture the sequential dependencies in both the direction and the opposite direction of the travel flow. 

We consider the direction of the travel flow first, i.e., from left to right. 
%
A GRU unit learns sequential correlations by maintaining a hidden state $h_{j}\in \mathbb{R}^{M}$ at position $j$, which can be regard as an accumulated information of the positions on the left of position $j$. Specifically, $h_{j}=\mathit{GRU}(x_j, h_{j-1})$, where $x_j$ is the input feature vector at position $j$ and $h_{j-1}$ is the hidden state at position $j-1$, i.e., the hidden state of the left position. 
More specifically, the GRU unit is composed of the following computations as shown in Equations ~\ref{eq:gru1}, ~\ref{eq:gru2}, ~\ref{eq:gru3}, and ~\ref{eq:gru4}. 
First, the GRU unit computes an update gate $z_{j}$ and reset gate $r_{j}$, respectively.  
%

Both gates are contributed to control how much information from the left hidden states should be considered in order to make the final similarity score estimation accurate. By doing this, it is possible to remember and forget left hidden states which are found to be relevant and irrelevant for the final similarity score estimation.
%
\begin{align}
&\mathbf{r}_{j}=\sigma\left(\mathbf{W}_{r} {x}_{j} + \mathbf{U}_{r} \mathbf{{h}}_{j-1}\right) \label{eq:gru1} \\
& \mathbf{z}_{j}=\sigma\left(\mathbf{W}_{z} {x}_{j}+ \mathbf{U}_{z} \mathbf{{h}}_{j-1}\right) \label{eq:gru2} \\
& \mathbf{\tilde{h}}_{j}=\phi\left(\mathbf{W}_h {x}_{j}+\mathbf{U}_h \left(\mathbf{{r}}_j \odot \mathbf{h}_{j-1}\right)\right) \label{eq:gru3} \\
& \mathbf{h}_{j}=\mathbf{z}_{j} \odot \mathbf{h}_{j}+\left(1-\mathbf{z}_{j}\right) \odot \mathbf{\tilde{h}}_{j} \label{eq:gru4}
\end{align}
where $\sigma$ is the logistic function, and $\odot$ denotes Hadamard product and $\phi$ is hyperbolic tangent function.
%
$x_j$ and $\mathbf{h}_j$ are the feature vector and hidden state at position $i$, respectively. 
$\mathbf{W}_{r}$, $\mathbf{W}_{z}$ $\mathbf{W}_{h}$, $\mathbf{U}_{r}$, $\mathbf{U}_{z}$ and $\mathbf{U}_{h}$ are parameters to be learned.  

For the opposite direction of the travel flow, i.e., from right to left, we apply another GRU to generate hidden state $\mathbf{h}^\prime_{j}=\mathit{GRU}^\prime(x_j, \mathbf{h}^\prime_{j+1})$. Here, the input consists of the feature vector at position $j$ and the hidden state at position $j+1$, i.e., the right hidden state. 

The final hidden state $H_i$ at position $i$ is the concatenation of the hidden states from both GRUs, i.e., $H_i=\mathbf{h}_{j} \oplus \mathbf{h}^\prime_{j}$ where $\oplus$ indicates the concatenation operation.

\subsubsection{Fully Connected Layer}

We stack all outputs from the BD-GRU units into a long feature vector $f_i=\langle H_1\oplus H_2\oplus \ldots \oplus H_Z\rangle$ where $\oplus$ indicates the concatenation operation. 
Then, we apply a fully connected layer with weight vector $W_{FC}\in\mathbb{R}^{|f_i|\times 1}$ to produce a single value $\hat{sim}_i=f_i^T W_{FC}$, as the estimated similarity for the competitive path $P^\prime_i$.
%
\subsubsection{Loss Function}

The loss function of the basic framework is shown in Equation~\ref{eq:loss1}. 
\begin{align}
\mathcal{L}(\textbf{W})=\frac{1}{|n|} \sum_{i=1}^{n}\left(\hat{\mathit{sim}}_i-\mathit{sim}_{i}\right)^{2} + \lambda\|\textbf{W}\|_{2}^{2}
\label{eq:loss1}
\end{align}


The first term of the loss function measures the discrepancy between the estimated similarity $\hat{sim}_i$ and the ground truth similarity $\mathit{sim}_i$. 
We use the average of square error to measure the discrepancy, where $n$ is the total number of competitive paths we used for training. 

The second term of the loss function is a L2 regularizer on all learnable parameters in the model, including the embedding matrix $B$, multiple matrices used in BD-GRU, and the matrix in the final fully connected layer $W_{FC}$. Here, $\lambda$ controls the relative importance of the second term w.r.t. the first term. The basic training pipeline is outlined in Algorithm~\ref{alg:BF}.
\begin{algorithm}
\caption{Training Pipeline of Basic \emph{PathRank}}
\label{alg:BF}
\LinesNumbered
\KwIn{Top-$k$ diversified path sequence: $\left(p_{1},p_{2},\ldots,
p_{n}\right) \in P$; Corresponding similarity between ground truth and top-$k$ diversified paths:
$\left(sim_{1},sim_{2},\ldots,sim_{n}\right) \in Sim$; Road network $G = (\mathbb{V}, \mathbb{E}, D, T, F)$} 
\KwOut{Driver driving preference similarity}
Use Node2vec on $G$ and get the embedding result $x_{emb}$\;
Initialize all learnable parameters $w$ in \emph{PathRank}\;
Initialize $MSE_{previous}^{traning}$ and $MSE_{previous}^{validation}$ \;
\Repeat{Maximum Epoch}{
 \textbf {R}andomly select a batch of instance $P_{bt}$ with $Sim_{bt}$ from $P$ and $Sim$\;
 \textbf {L}ooking up $x_{emb}$ for current node in path sequence, then inputting node vector to GRU\;
 \textbf {O}ptimize $w$ by minimizing the loss function Eq.(3) with $P_{bt}$ with $Sim_{bt}$ to learn \emph{PathRank} model\;
 \If{$MSE_{current}^{training}$ $<$ $MSE_{previous}^{traning}$}{
        Update $MSE_{previous}^{traning}$\;
        \If{$MSE_{current}^{validation}$ $<$ $MSE_{previous}^{validation}$}{
            Update $MSE_{previous}^{validation}$\;
            Saving training model\;
        }
  }
}
\end{algorithm}



\subsection{Advanced Framework}

\begin{figure*}
\includegraphics[width=\textwidth]{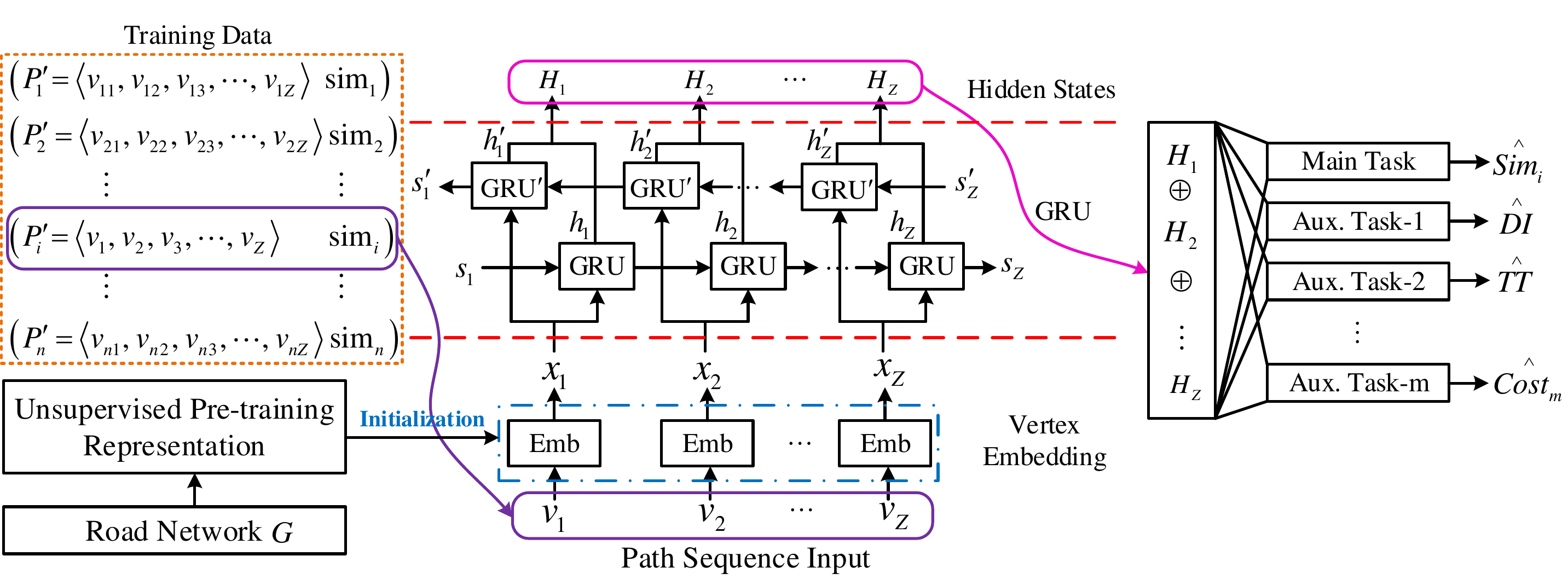}
  \caption{Advanced \emph{PathRank} Overview.}
  \label{fig:advandePR}
\end{figure*}

To further improve the learning accuracy, we pay particular attentions on the vertex embedding network since so far the vertex embedding network is ``graph-blind'' which only employs an embedding matrix $B$ and does not take into account any information from the underlying road network graph. 
To improve this, we design an advanced framework to extend the basic framework with the help of multi-task learning such that the embedding network takes into account both the topology of the underlying road network graph and the spatial properties associated with the underlying road network such as distances, travel times, and fuel consumption. 
The advanced \emph{PathRank} framework is shown in Figure~\ref{fig:advandePR}. 

\subsubsection{Capturing Graph Topology with Graph Embedding}
%
Graph embedding, e.g., DeepWalk~\cite{b17}, node2vec~\cite{b8}, LINE~\cite{b18}, GraphGAN~\cite{b21},  aims at learning low-dimensional, latent representations of vertices in a graph by taking into account the graph topology. %

A typical way to enable graph embedding is to mimic the way of embedding words for natural languages~\cite{b8,b17}. 
In particular, multiple vertex sequences can be generated by using random walks, where random walks can consider edge weights or ignore edge weights. 
Next, vertices are considered as words and the generated vertex sequences are considered as sentences, which enanles the use of word embedding techniques to generate embeddings for vertices. Since the vertex sequences are generated by applying random walks on the graph, the obtained vertex embedding actually already takes into account the graph topology. 

The learned vertex embeddings are used as feature vectors, which enables a wide variety of learning tasks on graphs such as classification~\cite{b17,b18}, link prediction~\cite{b28}, clustering~\cite{b18,b29}, recommendation~\cite{b30}, and visualization~\cite{b31,b32}. %

We propose two different strategies to incorporate graph embedding. First, we simply apply an existing graph embeding method, e.g., DeepWalk or node2vec, to embed a one-hot representation of a vertex to a low dimensional feature vector. Then, we use the feature vector as the input to the BD-GRU. This means that \emph{PathRank} only includes a RNN module, whose inputs are sequences of feature vectors, and the vertex embedding module is disabled.  

Second, inspired by the well-known practice of unsupervised pre-training~\cite{b33}, we use the embedding matrix obtained from an existing graph embedding method to initialize the embedding matrix $B$ in the vertex embedding module in \emph{PathRank}. 
This allows \emph{PathRank} to update the embedding matrix $B$ during training such that it not only captures the graph topology but also better fits the similarity regression.
\subsubsection{Capturing Spatial Properties with Multi-Task Learning}

Although many vertex embedding algorithms exist, they are only able to capture graph topology because they only focus on graphs representing, e.g., social networks and citation network. In other words, they do not consider graphs representing spatial networks such as road networks. 
However, in road network graphs, many spatial attributes, in addition to topology, are also very important. For example, distances between two vertices are crucial features for spatial networks. 
To let the graph embedding also maintain the spatial properties, we design a multi-task learning framework using pre-trained graph embedding. %

We first employ an existing graph embedding algorithm to initialize the vertex embedding matrix $B$ in the vertex embedding module of \emph{PathRank}. This pre-trained embedding matrix captures the graph topology. 
Next, we try to update $B$ such that it also captures relevant spatial properties during training. 
To this end, we employ multi-task learning principles, where the main task is to estimate similarity and the auxiliary tasks are to reconstruct travel costs of competitive paths which help learning an appropriate embedding matrix $B$ that also considers spatial properties of the underlying road network.

To enable the multi-task learning framework, in the final fully connected layer, we let \emph{PathRank} not only estimate a similarity score but also estimate, or reconstruct, the spatial properties of the corresponding competitive path $P^\prime_i$, such as the distance, travel time, and fuel consumption of $P^\prime_i$. %
We also extend the loss function to include terms that consider the discrepancies between the actual distance and the estimated distance, the actual travel time and the estimated travel time, and the actual fuel consumption and the estimated fuel consumption.
The loss function for the multi-task learning framework is defined in Equation~\ref{eq:loss2}. 
%
\begin{align}
\begin{split}\label{eq:loss2}
    \mathcal{L}(\textbf{W}) =&{}\frac{1}{|n|}[ \left(1-\alpha\right)\cdot \sum_{i=1}^{n}\left(\hat{\mathit{sim}}_i-\mathit{sim}_{i}\right)^{2} + \\
& \alpha \cdot \sum_{i=1}^{n} \sum_{k=1}^{m}\left(\hat {y}_{i}^{(k)}-y_{i}^{(k)}\right)^{2}] + \lambda\|\textbf{W}\|_{2}^{2}
\end{split}
\end{align}
where $\alpha$ is a hyper parameter that controls the trade-off between main task and auxiliary tasks; 
$\hat {y}_{i}^{(k)}$ and $y_{i}^{(k)}$ denote the estimated cost of the $k$-th auxiliary task and the ground truth of the $k$-th auxiliary task, respectively. For example, when considering distance, travel time, and fuel consumption, we set $m$ to 3; and $\hat {y}_{i}^{(k)}$ and $y_{i}^{(k)}$ represent the estimated and ground truth distance, travel time, or fuel consumption of the $i$-th competitive path $P^\prime_i$. The basic training pipeline is outlined in Algorithm~\ref{alg:AdF}.


\begin{algorithm}
\caption{Training Pipeline of Advanced \emph{PathRank}}
\label{alg:AdF}
\LinesNumbered
\KwIn{Top-$k$ diversified path sequence: $\left(p_{1},p_{2},\ldots,
p_{n}\right) \in P$; Corresponding similarity between ground truth and top-$k$ diversified paths:
$\left(sim_{1},sim_{2},\ldots,sim_{n}\right) \in Sim$; Auxiliary tasks, e.g., $\left(Cost_{TT},Cost_{Dis},Cost_{FC}\right)$; Road network $G = (\mathbb{V}, \mathbb{E}, D, T, F)$} 
\KwOut{Driver driving preference similarity, Auxiliary tasks prediction}
Use Node2vec on $G$ and get the network representation result $x_{emb}$\;
Initialize the embedding layer by using unsupervised pre-training representation\;
Initialize all learnable parameters $w$ in \emph{PathRank}\;
Initialize $MSE_{previous}^{traning}$ and $MSE_{previous}^{validation}$ \;
\Repeat{Maximum Epoch}{
 \textbf {R}andomly select a batch of instance $P_{bt}$ with $Sim_{bt}$ from $P$ with $Sim$\ and corresponding auxiliary tasks\;
 \textbf {L}ooking up $x_{emb}$ for current node in path sequence, then inputting node vector to GRU\;
 \textbf {O}ptimize $w$ by minimizing the loss function Eq.(8) with $P_{bt}$ with $Sim_{bt}$ and corresponding auxiliary tasks to learn \emph{PathRank} model and fine-tuning the network representation\;
 \If{$MSE_{current}^{training}$ $<$ $MSE_{previous}^{traning}$}{
        Update $MSE_{previous}^{traning}$\;
        \If{$MSE_{current}^{validation}$ $<$ $MSE_{previous}^{validation}$}{
            Update $MSE_{previous}^{validation}$\;
            Saving training model\;
        }
    }
}
\end{algorithm}

\section{Experiments}







We conduct a comprehensive empirical study to investigate the effectiveness of the proposed \emph{PathRank} framework. 

\subsection{Experiments Setup}

\subsubsection{Road Network and Trajectories} We consider the road network in North Jutland, Denmark. We obtain the road network graph from OpenStreetMap, which consists of 8,893 vertices and 10,045 edges. 

We use a substantial GPS data set occurred on the road network, which consists of 180 million GPS records for a two-year period from 183 vehicles. The sampling rate of the GPS data is 1 Hz (i.e., one GPS record per second). 
We split the GPS records into 22,612 trajectories representing different trips. 
A well-know map matching method~\cite{b22} is used to map match the GPS trajectories such that for each trajectory, we obtain its corresponding trajectory path. 

\subsubsection{Ground Truth Data} 

\begin{figure*}[htp]
     \centering
     \begin{subfigure}[b]{0.329\textwidth}
         \centering
         \includegraphics[width=\textwidth,height=3.6cm]{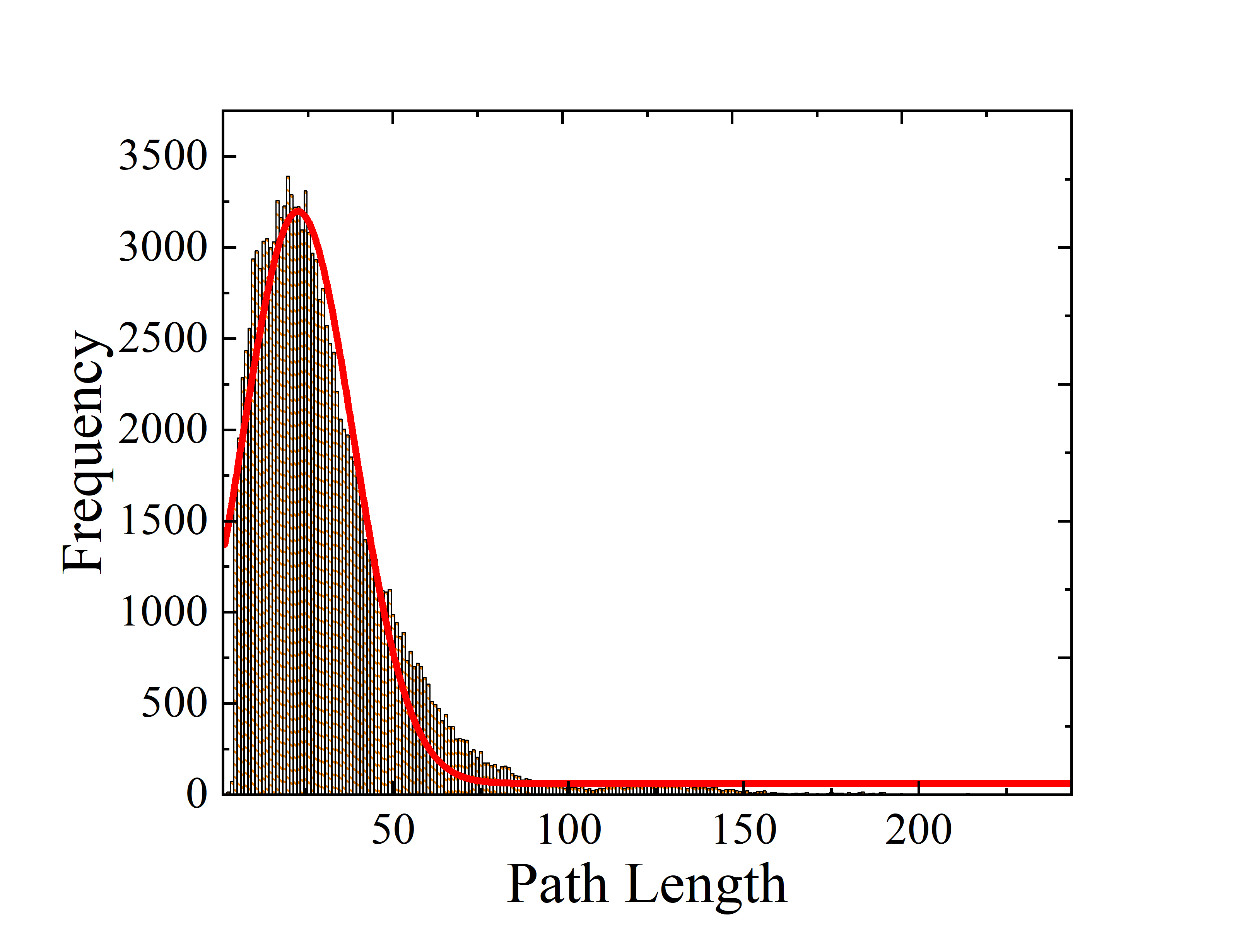}
         \caption{Training Set}
         \label{fig:subfig:train}
     \end{subfigure}
     \hfill
     \begin{subfigure}[b]{0.329\textwidth}
         \centering
         \includegraphics[width=\textwidth,height=3.6cm]{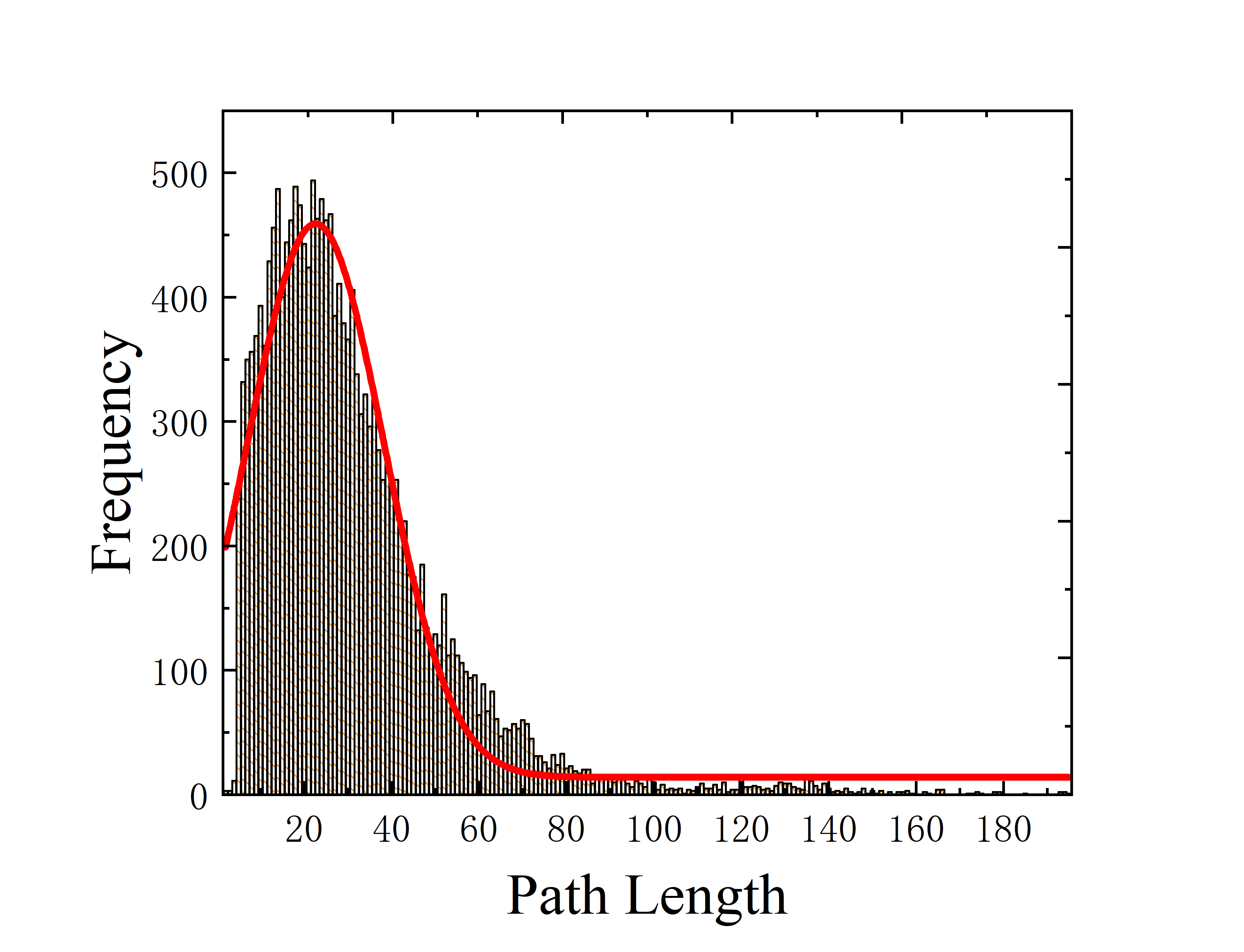}
         \caption{Validation Set}
         \label{fig:subfig:val}
     \end{subfigure}
     \hfill
     \begin{subfigure}[b]{0.329\textwidth}
         \centering
         \includegraphics[width=\textwidth,height=3.6cm]{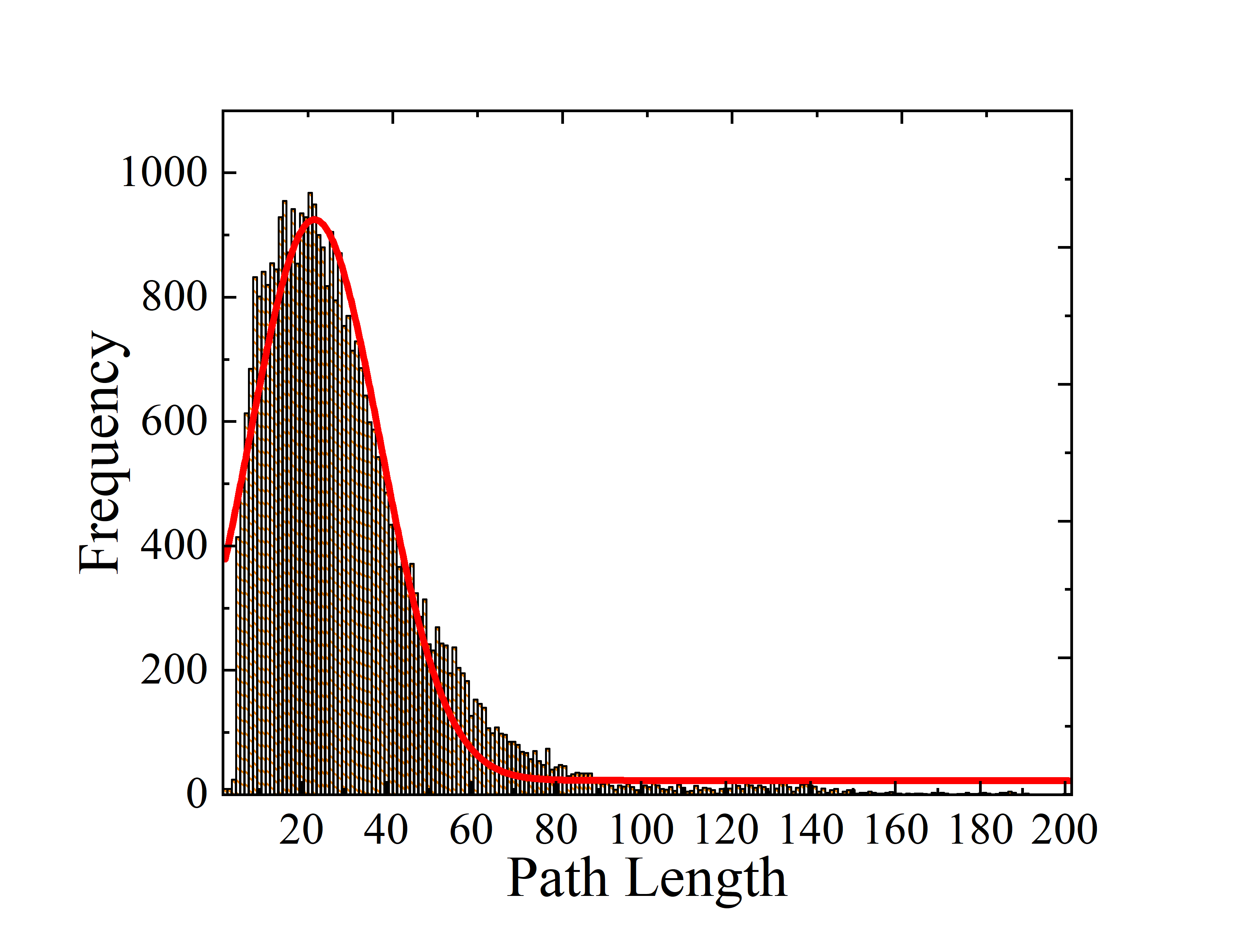}
         \caption{Testing Set}
         \label{fig:subfig:test}
     \end{subfigure}
        \caption{Cardinalities of the trajectory paths.}
        \label{fig:statistics}
        \vspace{-0pt}
\end{figure*}

We split the trajectories into three sets---70$\%$ for training, 10$\%$ for validation, and 20$\%$ for testing. The distributions of the cardinalities of the trajectory paths in training and testing sets are show in Figure~\ref{fig:statistics}. The distribution on the validation set is similar and thus is omitted due to space limitation. 

For each trajectory $T$, we obtain its source $s$, destination $d$, and the trajectory path $P_T$.  
Then, we employ seven different strategies to generate seven sets of competitive paths according to the source-destination pairs $(s$, $d)$---top-$k$ shortest paths (\emph{TkDI}), top-$k$ fastest paths (\emph{TkTT}), top-$k$ most fuel efficient paths (\emph{TkFC}), diversified top-$k$ shortest paths (\emph{D-TkDI}), diversified top-$k$ fastest paths (\emph{D-TkTT}), diversified top-$k$ most fuel efficient paths (\emph{D-TkFC}), and diversified, multi-cost top-$k$ paths (\emph{D-TkM}). For each competitive path $P$, we employ weighted Jaccard similarity $sim(P$, $P_T)$ as $P$'s \emph{ground truth} ranking score. 

When training and validation, we use the competitive path set generated by a specific training data generation strategy to train a \emph{PathRank} model. Thus, we are able to train seven different \emph{PathRank} models using the same set of training and validation trajectories, but seven different sets of competitive paths. 

When testing, to make the comparision among different \emph{PathRank} models fair, for each testing trajectory, we consider all competitive path sets generated by the 7 different strategies. This makes sure that (1) \emph{PathRank} models that are trained on different training data sets are tested against the same set of competitive paths; (2) a \emph{PathRank} model that is trained on a specific strategy is tested against competitive paths sets that are generated by all strategies. 


\subsubsection{PathRank Frameworks}
We consider different variations of \emph{PathRank}. 
First, we consider the basic framework \emph{PR-B} where the vertex embedding just employs an embedding matrix $B$, which ignores the graph topology. 

Second, we consider the advanced framework where the vertex embeding employs graph embedding. 
Recall that we have two strategies to use the advanced framework---keeping the graph embedding static (\emph{PR-A1}) vs. keep updating the embedding together with the \emph{PathRank} (\emph{PR-A2}). 

Finally, we consider the multi-task learning method which considers spatial properties, where we use \emph{PR-A2-M}$x$ to indicate a \emph{PathRank} model that uses an objective function considering $x$ spatial properties, i.e., $x$ auxiliary tasks. 

For the advanced frameworks, i.e., \emph{PR-A1}, \emph{PR-A2}, and \emph{PR-A2-M}$x$, we choose node2vec~\cite{b8} as the graph embeding method. 
%
Node2vec is a more general random walk based graph embedding method, which outperforms alternative methods such as DeepWalk~\cite{b17} and LINE~\cite{b18}. 
When new, better unsupervised graph embedding method becomes available, it can be easily integrated into \emph{PathRank} to replace node2vec. 

\subsubsection{Parameters}

When generating diversified top-$k$ paths, we consider two different similarity thresholds $\delta$---0.6 and 0.8. A smaller threshold enforces more diversified paths. However, it is also more likely that we cannot identify $k$ paths that are significantly diversified paths, especially when $k$ is large. 
Recall that the vertex embeding utilizes a embedding matrix $B \in \mathbb{R}^{M \times N}$ to embed each vertex into a $M$-dimensional feature vector, where $N$ is the number of vertices. 
We consider two settings of $M$, namely 64 and 128. 

We consider 250 GRU units in total in the bi-directional GRU module by considering the cardinalities of the trajectory paths shown in Figure~\ref{fig:statistics}, where the largest cardinality is 250.  
If a competitive path that consists of less than 250 vertices, we use zero padding to fill in. 

For the multi-task learning framework, we vary $\alpha$ from 0, 0.2, 0.4, 0.6, to 0.8 to study the effect on learning additional spatial properties. 


We summary different parameter settings in Table~\ref{tbl:Paramerters}, where the default values are shown in bold. 
\begin{table}[h]
\caption{Parameters of \emph{PathRank}}
\label{tbl:Paramerters}
\centering
\begin{tabular}{lc}
\toprule[2pt]
Parameters &Values \\ \midrule[1pt]
Similarity Threshold $\delta$  &0.6, \textbf{0.8} \\
Embedding feature size $M$            &64, \textbf{128} \\
Multi-task learning parameter $\alpha$       &{0}, 0.2, 0.4, 0.6, 0.8 \\
\bottomrule[2pt]
\end{tabular}
\end{table}


\subsubsection{Evaluation Metrics}

We evaluate the accuracy of the proposed \emph{PathRank} framework based on two categories of metrics. The first category includes metrics that measure how accurate the estimated ranking scores w.r.t. the ground truth ranking scores. This category includes Mean Absolute Error (MAE) and Mean Absolute Relative Error (MARE). Smaller MAE and MARE values indicate higher accuracy.  
Specifically, we have 
%
\begin{gather}
\operatorname{MAE}=\frac{1}{n} \sum_{i=1}^{n}\left|x_{i}-\hat{x}_{i}\right|;~~~
  \operatorname{MARE}=\frac{\sum_{i=1}^{n}\left|x_{i}-\hat{x}_{i}\right|}{\sum_{i=1}^{n}\left|x_{i}\right|}
\end{gather}
where $x_{i}$ and $\hat{x}_{i}$ represent the ground truth ranking score and the estimated ranking score, respectively; and $n$ is the total number of estimations. 
%


The second category includes Kendall rank correlation coefficient (denoted by $\tau$) and Spearman's rank correlation coefficient (denoted by $\rho$), which measure the similarity, or consistency, between a ranking based on the estimated ranking scores and a ranking based on the ground truth ranking scores. Sometimes, although the estimated ranking scores deviate from the ground truth ranking scores, the two rankings derived by both scores can be consistent. In this case, we consider the estimated ranking scores also accurate, since we eventually care the final rankings of the candidate paths but not the specific ranking scores for individiual candidate paths. 
Both $\tau$ and $\rho$ are able to measure how consistent between the two rankings. The higher the values are, the more consistent the two rankings are. If the two rankings are identical, both $\tau$ and $\rho$ values are 1. 
Specifically, we have 
\begin{gather}
\tau=\frac{{N}_{con} - {N}_{dis}}{n(n-1) / 2};~~~   
\rho=1-\frac{6\sum_{i=1}^n d_i^2}{n(n^2-1)} 
\end{gather}
Assume that we have a set of $n=3$ candidate paths $\{P_1$, $P_2$, $P_3\}$, the ground truth ranking is $\langle P_1$, $P_2$, $P_3\rangle$, and the estiamted ranking is $\langle P_2$, $P_3$, $P_1\rangle$.

In $\tau$, ${N}_{con}$ and ${N}_{dis}$ represent the number of path pairs are consistent and inconsistent in the two rankings. We have ${N}_{con}=1$ since in both ranking, $P_2$ appears before $P_3$. We have ${N}_{dis}=2$ since $P_1$ appears before $P_3$ in the ground truth ranking, while $P_3$ appears before $P_1$ in the estimated ranking. Similarly, the orderings between $P_1$ and $P_2$ are also inconsistent in two rankings.  

In $\rho$, $d_i$ represents the rank difference on the $i$-th competitive path in both rankings. 
Following the running example, we have $d_1=1-3=-2$ because path $P_1$ has rank 1 and rank 3 in both rankings, respectively.

\subsection{Experimental Results}
\subsubsection{Effects of Training Data Generation Strategies}

We investigate how the different training data generation strategies affect the accuracy of \emph{PathRank}. 
We first consider PR-A1, where we only use graph embedding method node2vec to initialize the vertex embeding matrix $B$ and do not update $B$ during training. 
%

%

Table~\ref{tbl:1} shows the results, where we categorize the training data generation strategies into three categories based on top-$k$ paths, diversified top-$k$ paths, and multi-cost, diversified top-$k$ paths. For each category, the best results are highlighted with underline. The best results overall is also highlighted with bold. 
We also show results when the embeding feature sizes are $M=64$ and $M=128$, respectively.

The results show that (1) when using the diversified top-$k$ paths for training, we have higher accuracy (i.e., lower MAE and MARE and larger $\tau$ and $\rho$) compared to when using top-$k$ paths; (2) using multi-cost, diversified top-$k$ paths achieves better accuracy compared to single-cost, diversified top-$k$ paths, thus achieving the best results; (3) a larger embeding feature size $M$ achieves better results. 

\begin{table}[h]
\caption{Training Data Generation Strategies, \emph{PR-A1}}
\label{tbl:1}
\centering
\begin{tabular}{cccccc}
\toprule[2pt]
\textbf{Strategies} &\textbf{$M$} &\textbf{MAE} &\textbf{MARE} &\textbf{$\tau$} &$\rho$ \\
\midrule[1pt]
\multirow{2}*{\emph{TkDI}}   &64       &0.1433       &0.2300        &0.6638               &0.7044 \\
                             &128      &\underline{0.1168}       &\underline{0.1875}        &\underline{0.6913}               &\underline{0.7330} \\
\multirow{2}*{\emph{TkTT}}   &64       &0.1302       &0.2090        &0.6642               &0.7046 \\
                             &128      &0.1181       &0.1896        &0.6818               &0.7208 \\ 
\multirow{2}*{\emph{TkFC}}   &64       &0.1208       &0.1940        &0.6692               &0.7131 \\
                             &128      &0.1257       &0.2019        &0.6699               &0.7110 \\  \midrule[1pt]
\multirow{2}*{\emph{D-TkDI}} &64       &0.1140       &0.1830        &0.6959               &0.7346 \\
                             &128      &0.0955       &0.1533        &0.7077               &0.7492 \\
\multirow{2}*{\emph{D-TkTT}} &64       &0.1050       &0.1686        &0.7124               &0.7554 \\
                             &128      &0.0974       &0.1564        &0.7271               &\underline{0.7714} \\ 
\multirow{2}*{\emph{D-TkFC}} &64       &0.1045       &0.1678        &0.7100               &0.7544 \\
                             &128      &\underline{0.0900}       &\underline{0.1445}        &\underline{0.7238}               &0.7685 \\ \midrule[1pt]
\multirow{2}*{\emph{D-TkM}}  &64       &0.1077       &0.1729        &0.7261               &0.7679 \\
                             &128      &\underline{\textbf{0.0792}}       &\underline{\textbf{0.1271}}        &\underline{\textbf{0.7478}}               &\underline{\textbf{0.7876}} \\ 
\bottomrule[2pt]
\end{tabular}
\end{table}

Next, we consider PR-A2, where the graph embedding matrix $B$ is also updated during training to fit better the ranking score regression problem.  Table~\ref{tbl:2} shows the results. The three observations from Table~\ref{tbl:1} also hold for Table~\ref{tbl:2}. In addition, PR-A2 achieves better accuracy than does \emph{PR-A1}, meaning that updating embedding matrix $B$ is useful.

%

\begin{table}[h]
\caption{Training Data Generation Strategies, \emph{PR-A2}}
\label{tbl:2}
\centering
\begin{tabular}{cccccc}
\toprule[2pt]
\textbf{Strategies} &\textbf{$M$} &\textbf{MAE} &\textbf{MARE} &\textbf{$\tau$} &\textbf{$\rho$} \\
\midrule[1pt]
\multirow{2}*{\emph{TkDI}}   &64       &0.1163       &0.1868        &0.6835               &0.7256 \\
                             &128      &0.1130       &0.1814        &\underline{0.7082}               &\underline{0.7481} \\
\multirow{2}*{\emph{TkTT}}   &64       &0.1218       &0.1956        &0.6858               &0.7282 \\
                             &128      &0.1161       &0.1864        &0.7026               &0.7446 \\ 
\multirow{2}*{\emph{TkFC}}   &64       &0.1216       &0.1952        &0.6911               &0.7321 \\
                             &128      &\underline{0.1082}       &\underline{0.1737}        &0.7070               &0.7477 \\ \midrule[1pt]
\multirow{2}*{\emph{D-TkDI}} &64       &0.0940       &0.1509        &0.7144               &0.7532 \\
                             &128      &0.0855       &0.1373        &0.7339               &0.7731 \\
\multirow{2}*{\emph{D-TkTT}} &64       &0.1010       &0.1622        &0.7283               &0.7693 \\
                             &128      &0.0997       &0.1600        &0.7169               &0.7596 \\ 
\multirow{2}*{\emph{D-TkFC}} &64       &0.0938       &0.1506        &0.7318               &0.7743 \\
                             &128      &\underline{0.0809}       &\underline{0.1299}        &\underline{0.7386}               &\underline{0.7811} \\ \midrule[1pt]
\multirow{2}*{\emph{D-TkM}}  &64       &0.0966       &0.1551        &0.7393               &0.7771 \\
                             &128      &\underline{\textbf{0.0725}}       &\underline{\textbf{0.1164}}        &\underline{\textbf{0.7528}}               &\underline{\textbf{0.7905}} \\ 
\bottomrule[2pt]
\end{tabular}
\end{table}


From the above experiments, the multi-cost, diversifed top-$k$ strategy \emph{D-TkM} is the most promising strategy. Then, we further investigate the effects on the similarity threshold $\delta$ used in the diversifed top-$k$ path finding. Specifically, we consider two threshold values 0.6 and 0.8 and the results are shown in Table~\ref{tbl:3Diversification}. 
When a smaller threshold is used, i.e., higher diversity in the top-$k$ paths, the accuracy is improved.

\begin{table}[h]
\caption{Effects of Similarity Threshold $\delta$}
\label{tbl:3Diversification}
\centering
\begin{tabular}{ccccccc}
\toprule[2pt]
&$\delta$
&\textbf{$M$} &\textbf{MAE} &\textbf{MARE} &\textbf{$\tau$} &\textbf{$\rho$} \\
\midrule[1pt]
\multirow{4}*{\emph{PR-A1}} &\multirow{2}*{0.6}  &64       &0.1006       &0.1615        &0.7321               &0.7733 \\
                   &                      &128      &\underline{0.0770}       &\underline{0.1237}        &\underline{0.7496}               &0.7874 \\ 
                   &\multirow{2}*{0.8}    &64       &0.1077       &0.1729        &0.7261               &0.7679 \\
                   &                      &128      &0.0792       &0.1271        &0.7478               &\underline{0.7876} \\  \midrule[1pt]
\multirow{4}*{\emph{PR-A2}}  &\multirow{2}*{0.6} &64       &0.0817       &0.1311        &0.7404               &0.7792 \\
                   &                      &128      &\underline{\textbf{0.0710}}       &\underline{\textbf{0.1140}}        &\underline{\textbf{0.7751}}               &\underline{\textbf{0.8109}} \\ 
                   &\multirow{2}*{0.8}    &64       &0.0966       &0.1551        &0.7393               &0.7771 \\
                   &                      &128      &0.0725       &0.1164        &0.7528               &0.7905 \\ 
\bottomrule[2pt]
\end{tabular}
\vspace{-10pt}
\end{table}



Since we have identified that \emph{D-TkM} gives the best accuracy, we only consider \emph{D-TkM} with similarity threshold $\delta=0.8$ in the diversified top-$k$ paths as the training data generation strategy for the following experiments. 

\subsubsection{Effects of Vertex Embedding}

We investigate the effects of different vertex embedding strategies. We consider PR-B where we just use a randomly initialized embeding matrix $B$, which totally ignores graph topology. 
For PR-A1 and PR-A2 where we both use node2vec to embed vertices. Here, we use node2vec to embed both weighted and unweighted graphs, respectively. When embeding weighted graphs, we simply use distance as edge weights. 

Based on the results in Table~\ref{tbl:embeddings}, we observe the following. First, \emph{PR-B} gives the worst accuracy: the estimated ranking scores have the largest errors in terms of both MAE and MARE; and the ranking based on estiamted ranking scores deviates the most from the ground truth ranking in terms of both $\tau$ and $\rho$. This suggests that ignoring graph topology when embeding vertices is not a good choice. 

Second, when embeding vertices using node2vec, whether or not considering edge weights does not significantly change the accuracy. Thus, it is not a significant design choice 

Third, PR-A2 achieves the best accuracy in terms of both errors on estimated ranking scores and consistency between two rankings. Thus, this suggests that considering graph topology improves accuracy and updating the embeding matrix $B$ according to the loss function on ranking scores makes the embedding matrix fit better the ranking score regression problem. This also suggests that, by including spatial properties in the loss function, the embeding matrix $B$ should be tuned to capture spatial properties, which in turn should improve ranking score regression. This is verified in the following experiments on the multi-task framework.  



\begin{table}[h]
\caption{Effects of Embedding}
\label{tbl:embeddings}
\centering
\begin{tabular}{cccccc}
\toprule[2pt]
&\textbf{Embedding} &\textbf{MAE} &\textbf{MARE} &\textbf{$\tau$} &\textbf{$\rho$} \\ \midrule[1pt]
\emph{PR-B}                &---           &\underline{0.1159}       &\underline{0.1816}        &\underline{0.7233}               &\underline{0.7611} \\ \midrule[1pt]   
\multirow{2}*{\emph{PR-A1}} &unweighted    &0.0878       &0.1410        &0.7453               &0.7852 \\
                       &weighted      &\underline{0.0792}       &\underline{0.1271}        &\underline{0.7478}               &\underline{0.7876} \\ \midrule[1pt]
\multirow{2}*{\emph{PR-A2}} &unweighted    &0.0734       &0.1178        &\underline{\textbf{0.7640}}               &\underline{\textbf{0.8012}} \\
                       &weighted      &\underline{\textbf{0.0725}}       &\underline{\textbf{0.1164}}        &0.7528               &0.7905 \\ 
\bottomrule[2pt]
\end{tabular}
\end{table}

\subsubsection{Effects of Multi-task Learning}

In the following set of experiments, we study the effects of the proposed multi-task learning framework. 
In particular, we investigate how much we are able to improve when incorporating different spatial properties in the loss function to let the vertex embedding also consider spatial properties, which may potentially contribute to better ranking score regression. 


We start by \emph{PR-A2-M1}, which considers only one auxiliary task on reconstructing distances. 
This means that \emph{PathRank} not only estimate the ranking score of a competitive path but also tries to reconstruct the distance of the competitive paths. 
Table~\ref{tb1:multitask} shows the results with varying $\alpha$ values. 
When $\alpha=0$, the auxiliary task is ignored, which makes \emph{PR-A2-M1} into \emph{PR-A2}, i.e., its corresponding model with only the main task on estimating ranking scores. 
When $\alpha>0$, i.e., the auxiliary task on distances is considered while learning, we observe that the estimated ranking scores are improved. 
In particular, the setting with $\alpha = 0.6$ gives the best results in terms both $\tau$ and $\rho$, indicating that the ranking w.r.t. the estimated ranking scores is more consistent with the ground truth ranking. 
When $\alpha = 0.8$, it achieves the smallest MAE and MARE. Both settings suggest that considering the additional auxiliary task on reconstructing distance helps improve the final ranking. 

\emph{PR-A2-M2} includes two auxiliary tasks on reconstructing both distances and travel times, and \emph{PR-A2-M3} includes three auxiliary tasks on reconstructing distances, travel times, and fuel consumption. All the three multi-task models show that considering spatial properties improve the final ranking. In particular, when considering all the three spatial properties give the best final ranking in terms of $\tau$ and $\rho$, i.e., achieving the most consistent ranking w.r.t. the ground truth ranking. 




\begin{table}[ht]
\caption{Effects of $\alpha$, \emph{PR-A2-Mx}}
\label{tb1:multitask}
\centering
\begin{tabular}{cccccc}
\toprule[2pt]
&\textbf{$\alpha$} &\textbf{MAE} &\textbf{MARE} &\textbf{$\tau$} &\textbf{$\rho$} \\ \midrule[1pt]
\emph{PR-A2}                    &0        &\underline{0.0725}       &\underline{0.1164}        &\underline{0.7528}               &\underline{0.7905} \\ \midrule[1pt]
\multirow{4}*{\emph{PR-A2-M1}}  &0.2      &0.0756       &0.1214        &0.7713               &0.8057 \\
                                &0.4      &0.0704       &0.1129        &0.7765               &0.8110 \\
                                &0.6      &0.0693       &0.1113        &\underline{0.7783}               &\underline{0.8141} \\
                                &0.8      &\underline{0.0680}       &\underline{\textbf{0.1029}}        &0.7712               &0.8057 \\ \midrule[1pt]
\multirow{4}*{\emph{PR-A2-M2}}  &0.2      &\underline{\textbf{0.0653}}       &\underline{0.1048}        &0.7727               &0.8089 \\
                                &0.4      &0.0701       &0.1125        &\underline{0.7869}               &\underline{0.8235} \\
                                &0.6      &0.0777       &0.1247        &0.7752               &0.8100 \\
                                &0.8      &0.0807       &0.1296        &0.7616               &0.7973 \\ \midrule[1pt]
\multirow{4}*{\emph{PR-A2-M3}}  &0.2      &0.0724       &0.1162        &0.7732               &0.8092\\
                                &0.4      &0.0740       &0.1188        &0.7711               &0.8090 \\
                                &0.6      &\underline{0.0662}       &\underline{0.1063}        &\underline{\textbf{0.7923}}               &\underline{\textbf{0.8261}} \\
                                &0.8      &0.0695       &0.1116        &0.7842               &0.8177 \\
\bottomrule[2pt]
\end{tabular}
\vspace{-10pt}
\end{table}

\subsection{Comparison with Baseline Ranking Heuristics}

We consider three baseline ranking heuristics, i.e., ranking the candidate paths according to their distances, travel times, and fuel consumption. When using each heuristics, we obtain a ranking. Then, we compare the ranking with the ground truth ranking to compute the corresponding $\tau$ and $\rho$. 

Table~\ref{tb1:finer} shows the comparision, where we categorize the testing cases based on the distances of the lengths of their corresponding trajectory paths into three categories (0, 5], (5, 10], and (10, 15] km. 
The results show that the ranking obtained by \emph{PathRank}, more specifically, by \emph{PR-A2-M3}, is clearly the best in all categories, suggesting that \emph{PathRank} outperforms baseline heuristics. 
In the longest distance category, \emph{PathRank} becomes less accurate since the most of the training paths are within short distance categories, as shown in Figure~\ref{fig:statistics}. 


\begin{table}[h]
\caption{Comparison with Baseline Ranking Heuristics}
\label{tb1:finer}
\centering
\begin{tabular}{lcccccc}
\toprule[2pt]
&\multicolumn{2}{c}{(0, 5{]}} & \multicolumn{2}{c}{(5, 10{]}} & \multicolumn{2}{c}{(10, 15{]}} \\ \cline{2-7}
 &$\tau$  &$\rho$        &$\tau$  &$\rho$       &$\tau$  &$\rho$ \\ \toprule[1pt]

\emph{Distance} &0.7569 &0.7886 &0.6562 &0.6912 &0.4745 &0.4361  \\
\emph{Travel Time} &0.6760 &0.7066 &0.6406 &0.6784 &0.4714 &0.5435  \\
\emph{Fuel} &0.6890 &0.7229 &0.3919 &0.4099 &0.2591 &0.2291  \\
\emph{PathRank}  &\textbf{0.7985} &\textbf{0.8334} &\textbf{0.7649} &\textbf{0.8055} &\textbf{0.6097} &\textbf{0.6702} \\
\bottomrule[2pt]
\end{tabular}
\vspace{-10pt}
\end{table}


\subsection{Comparison with Driver Specific PathRank}
\label{sec:exp-driver-specific}

We investigate if driver specific \emph{PathRank} models are able to provide more accurate, personalized ranking. 
We select the two drivers with the largest amount training trajectories. The Driver 1 has 2068 trajectories and Driver 2 has 1457 trajectories. 
We train three \emph{PR-A2-M3} models, denoted as \emph{PR-Dr1}, \emph{PR-Dr2}, and \emph{PR-All}, using the training trajectories from Driver 1, Driver 2, and all drivers, respectively.  

We test the three models using the testing trajectories from Driver 1 and Driver 2, respectively. Table~\ref{tb1:Driver} shows that (1) for the testing trajectories from Driver 1, \emph{PR-Dr1} outperforms \emph{PR-Dr2}; and for the testing trajectories from Driver 2, \emph{PR-Dr2} outperforms \emph{PR-Dr1}; (2) for both testing cases, \emph{PR-All} performs the best. 

\begin{table}[h]
\caption{Comparison with Driver Specific PathRank}
\label{tb1:Driver}
\centering
\begin{tabular}{clccccc}
\toprule[2pt]
\begin{tabular}[c]{@{}c@{}} \textbf{Testing} \\ \textbf{Data}\end{tabular}  &\textbf{Model} &\textbf{MAE} &\textbf{MARE} &\textbf{$\tau$} &\textbf{$\rho$} \\
\midrule[1pt]
\multirow{3}*{Driver1} &\emph{PR-Dr1}     &0.1154       &0.1878       &0.7868              &0.8162 \\
                       &\emph{PR-Dr2}     &0.1464       &0.2289       &0.7753               &0.7983 \\ 
                       &\emph{PR-All}       &\underline{\textbf{0.0614}}       &\underline{0.1000}       &\underline{0.8269}               &\underline{0.8560} \\ \midrule[1pt]
\multirow{3}*{Driver2} &\emph{PR-Dr1}     &0.2431      &0.3957        &0.6628               &0.6547 \\
                       &\emph{PR-Dr2}     &0.1066       &0.1666        &0.7825               &0.8037 \\ 
                       &\emph{PR-All}       &\underline{0.0633}       &\underline{\textbf{0.0989}}        &\underline{\textbf{0.8430}}               &\underline{\textbf{0.8610}} \\ 
\bottomrule[2pt]
\end{tabular}
\end{table}

Next, we report statistics on a case-by-case comparision, where  
Table~\ref{tb1:percentage} shows the percentages of the cases where a driver specific \emph{PathRank} outperforms \emph{PR-All}. 
Specifically, \emph{PR-Dr1} outperforms \emph{PR-All} in ca. 21\% of the testing cases from Driver 1, and \emph{PR-Dr2} outperforms \emph{PR-All} in ca. 17\% of the testing cases from Driver 2. 

\begin{table}[h]
\caption{Percentage when \emph{PR-Dr} Outperforms \emph{PR-All}}
\label{tb1:percentage}
\centering
\begin{tabular}{ccccc}
\toprule[2pt]
         & \multicolumn{2}{c}{\emph{PR-Dr1}} & \multicolumn{2}{c}{\emph{PR-Dr2}} \\ \cline{2-5}
         & $\tau$          &$\rho$          & $\tau$          & $\rho$          \\ \midrule[1pt]
 &21.57\%    &20.10\%        &17.24\%        &17.24\%              \\
\bottomrule[2pt]
\end{tabular}
\end{table}

The results from the above two tables suggest that user-specific \emph{PathRank} models have a potential to achieve personalized ranking, which may outperform the \emph{PathRank} model trained on all trajectories, i.e., \emph{PR-All}.  
However, the number of an individual driver's training trajectories is often very limited, making it difficult to cover a large feature space. Thus, it is often difficult to outperform \emph{PR-All} on average. 

\subsection{Online Efficiency}
Since ranking candidate paths is conducted online, we report the runtime. Table~\ref{tb1:testtime} reports the runtime for estimating a path when using different \emph{PathRank} models. It shows that the non-multi-task learning models, i.e., \emph{PR-B}, \emph{PR-A1}, and \emph{PR-A2}, have similar run time. Multi-task learning models take longer time and the more auxiliary tasks are included in a model, the longer time the model takes. 
\emph{PR-A2-M3} takes the longest time, on average 45.1 ms. Suppose that an advanced routing algorithm or a commercial navigation system returns 10 candidate paths, \emph{PR-A2-M3} is able to return a ranking in 451 ms, which is within a reasonable response time. 
\begin{table}[h]
\caption{Average Testing Runtime Per Path (ms)}
\label{tb1:testtime}
\centering
\begin{tabular}{cccccc}
\toprule[2pt]
\emph{PR-B}     &\emph{PR-A1} &\emph{PR-A2} &\emph{PR-A2-M1}     &\emph{PR-A2-M2}  &\emph{PR-A2-M3}\\ \midrule[1pt]
11.4           &11.3          &11.5          &22.8               &34.4             &45.1\\
\bottomrule[2pt]
\end{tabular}
\vspace{-10pt}
\end{table}

\subsection{Scalability}
We conduct this experiment to investigate the performance when varying the sizes of training data. Specifically, we use 25\%, 50\%, 75\%, 100\% of the total training data to train \emph{PathRank}, respectively. Based on the results shown in Table~\ref{tbl:sca}, more training data gives better performance. 


\begin{table}[h]
\caption{Effectiveness of the Size of Training Data}
\label{tbl:sca}
\centering
\begin{tabular}{ccccc}
\toprule[2pt]
\textbf{Percentage}  &\textbf{MAE} &\textbf{MARE} &\textbf{$\tau$} &\textbf{$\rho$} \\
\midrule[1pt]
25\%    &0.1260       &0.2023        &0.7100               &0.7535 \\
50\%    &0.1001       &0.1607        &0.7286              &0.7686 \\
75\%    &0.0830       &0.1333        &0.7395               &0.7795 \\
100\%   &\textbf{0.0725}      &\textbf{0.1164}        &\textbf{0.7528}              &\textbf{0.7905} \\
\bottomrule[2pt]
\end{tabular}
\end{table}

\section{Conclusion and Future work}

We propose \emph{PathRank}, a learning to rank technique for ranking paths in spatial networks. 
We propose an effective way to generate a compact set of competitive paths to enable effective and efficient learning. Then, we propose a multi-task learning framework to enable graph embedding that takes into account spatial properties. A recurrent neural network, together with the learned graph embedding, is employed to estimate the ranking scores which eventually enable ranking paths. Empirical studies conducted on a large real world trajectory set demonstrate that \emph{PathRank} is effective and efficient for practical usage. 
As future work, it is of interest to exploit an attention mechanism on path lengths to further improve the ranking quality of  \emph{PathRank}.


\small
\begin{thebibliography}{00}

\bibitem{b1} C. Guo, B. Yang, J. Hu, and C. S. Jensen, ``Learning to route with sparse trajectory sets," \textit{in ICDE}, 2018, pp. 1073-1084.
\bibitem{b2} Z. Ding, B. Yang, Y. Chi, and L. Guo, ``Enabling smart transportation systems: A parallel spatio-temporal database approach," \textit{IEEE Trans. Computers}, vol. 65, no. 5, pp. 1377-1391, 2016.
\bibitem{b3} V. Ceikute and C. S. Jensen. ``Routing service quality - local driver behavior versus routing services," \textit{in MDM}, 2013, pp. 97-106.
\bibitem{b4} B. Yang, C. Cuo, C. S. Jensen, M. Kaul and S. Shang, ``Stochastic skyline route planning under time-varying uncertainty," \textit{in ICDE}, 2014, pp. 136-147.
\bibitem{b5} J. Y. Yen, ``Finding the k shortest loopless paths in a network," \textit{Management Science}, vol. 17, no. 11, pp. 712-716, 1971.
\bibitem{b6} H. Liu, C. Jin, B. Yang and A. Zhou, ``Finding top-k shortest paths with diversity," \textit{IEEE Trans. Knowl. Data Eng.}, vol. 30, no. 3, pp. 488-502, 2018.
\bibitem{b7} B. Yang, C. Guo, Y. Ma and C. S. Jensen, ``Toward personalized, context-aware routing," \textit{The VLDB Journal}, vol. 24, no. 2, pp. 297-318, 2015.
\bibitem{b8} A. Grover and J. Leskovec, ``node2vec: Scalable feature learning for networks," \textit{in SIGKDD}, 2016, pp. 855-864.
\bibitem{b9} F. C. Grey, ``Inferring probability of relevance using the method of logistic regression," \textit{in SIGIR}, 1994, pp. 222-231.
\bibitem{b10} L. Rigutini, T. Papini, M. Maggini and F. Scarselli, ``Learning to rank by a neural-based sorting algorithm," \textit{in SIGIR}, 2008.
\bibitem{b11} T. Joachims, ``Optimizing search engines using clickthrough data," \textit{in SIGKDD}, 2002, pp. 133-142.
\bibitem{b12} Z. Cao, T. Qin, T. Liu, M. Tsai and H. Li, ``Learning to rank: from pairwise approach to listwise approach," \textit{in ICML}, 2007, pp. 129-136.
\bibitem{b13} P. Huang, X. He, J. Gao, L. Deng, A. Acero and L. Heck, `` Learning deep structured semantic models for web search using clickthrough data," \textit{in CIMK}, 2013, pp. 2333-2338.
\bibitem{b14} Y. Shen, X. He, J. Gao, L. Deng and G. Msenil, ``Learning semantic representations using convolutional nueral networks for web search," \textit{in WWW}, 2014, pp. 373-374.
\bibitem{b15} L. Pang, Y. Lan, J. Guo, J. Xu, J. Xu and X. Cheng, ``DeepRank: A new deep architecture for relevance ranking in information retrieval," \textit{in CIKM}, 2017, pp. 257-266.
\bibitem{b16} P. Cui, X. Wang, J. Pei, and W. Zhu, ``A survey on network embedding," \textit{IEEE Trans. Knowl. Data Eng.}, 2018, DOI: 10.1109/TKDE.2018.2849727.
\bibitem{b17} B. Perozzi, R. Ai-Rfou and S. Skjena, ``Deepwalk: Oneline learning of social representations," \textit{in SIGKDD}, 2014, pp. 701-710.
\bibitem{b18} J. Tang, M. Qu, M. Wang, M. Zhang, J. Yan and Q. Mei, ``Line: Large-scale information network embedding," \textit{in WWW}, 2015, pp. 1067-1077.
\bibitem{b19} T. Mikolv, C. Kar, C. Greg and D. Jeffrey, ``Efficient estimation  of word representations in vector space," \textit{arXiv preprint arXiv:1301.3781}, 2013.
\bibitem{b20} S. Cao, W. Lu, and Q. Xu, ``Deep neural networks for learning graph representation," \textit{in AAAI}, 2016.

\bibitem{DBLP:conf/icde/DaiYGD15}
J.~Dai, B.~Yang, C.~Guo, and Z.~Ding.
\newblock Personalized route recommendation using big trajectory data.
\newblock In {\em {ICDE}}, pages 543--554, 2015.

\bibitem{DBLP:journals/pvldb/DaiYGJH16}
J.~Dai, B.~Yang, C.~Guo, C.~S. Jensen, and J.~Hu.
\newblock Path cost distribution estimation using trajectory data.
\newblock {\em {PVLDB}}, 10(3):85--96, 2016.

\bibitem{DBLP:conf/gis/GuoM0JK12}
C.~Guo, Y.~Ma, B.~Yang, C.~S. Jensen, and M.~Kaul.
\newblock Ecomark: evaluating models of vehicular environmental impact.
\newblock In {\em {SIGSPATIAL}}, pages 269--278, 2012.

\bibitem{DBLP:journals/geoinformatica/Guo0AJT15}
C.~Guo, B.~Yang, O.~Andersen, C.~S. Jensen, and K.~Torp.
\newblock Ecomark 2.0: empowering eco-routing with vehicular environmental
models and actual vehicle fuel consumption data.
\newblock {\em GeoInformatica}, 19(3):567--599, 2015.


\bibitem{b21} H. Wang, J. Wang, J. Wang, M. Zhao, W. Zhang, F. Zhang, X. Xie, and M. Guo, ``GraphGAN: Graph representation learning with generative adversarial nets," \textit{in AAAI}, 2018, pp. 2508-2515.
\bibitem{b22} P. Newson and J. Krumm, ``Hidden markov map matching through noise and sparseness," \textit{in SIGSPATIAL}, 2009, pp. 336-343.
\bibitem{b23} X. Li, K. Zhao, G. Cong, C. S. Jensen and W. Wei, ``Deep representation learning for trajectory similarity computation," \textit{in ICDE}, 2018, pp. 617-628.
\bibitem{b24} J. Y. Yen, ``Finding the k shortest loopless paths in a network," \textit{Informs}, vol. 17, no. 11, pp. 712-716, 1971.
\bibitem{b25} J. Hershberger, M. Maxel and S. Suri, ``Finding the k shortest simple paths: A new algorithm and its implementation," \textit{ACM Trans. Algori.}, vol. 3, no. 4, pp. 45, 2007.
\bibitem{b26} N. Katoh, T. Ibaraki and H. Mine, ``An efficient algorithm for k shortest simple paths," \textit{Networks}, vol. 12, no. 4, pp. 411-427, 1982.
\bibitem{b27} D. Eppsetin, ``Finding the k shortest paths," \textit{SIAM Jour. Comput.}, vol. 28, no. 2, pp. 652-673, 1998.
\bibitem{DBLP:conf/icde/Guo0HJ18}
C.~Guo, B.~Yang, J.~Hu, and C.~S. Jensen.
\newblock Learning to route with sparse trajectory sets.
\newblock In {\em {ICDE}}, pages 1073--1084, 2018.

\bibitem{DBLP:journals/vldb/HuYGJ18}
J.~Hu, B.~Yang, C.~Guo, and C.~S. Jensen.
\newblock Risk-aware path selection with time-varying, uncertain travel costs:
a time series approach.
\newblock {\em {VLDB} J.}, 27(2):179--200, 2018.

\bibitem{DBLP:journals/geoinformatica/HuYJM17}
J.~Hu, B.~Yang, C.~S. Jensen, and Y.~Ma.
\newblock Enabling time-dependent uncertain eco-weights for road networks.
\newblock {\em GeoInformatica}, 21(1):57--88, 2017.

\bibitem{DBLP:conf/cikm/Kieu0GJ18}
T.~Kieu, B.~Yang, C.~Guo, and C.~S. Jensen.
\newblock Distinguishing trajectories from different drivers using incompletely
labeled trajectories.
\newblock In {\em {CIKM}}, pages 863--872, 2018.

\bibitem{IJCAI}
T.~Kieu, B.~Yang, C.~Guo, and C.~S. Jensen.
\newblock Outlier detection for time series with recurrent autoencoder
ensembles.
\newblock In {\em {IJCAI}}, 2019.


\bibitem{b28} D. Liben-Nowell and J. Kleinberg, ``The link-prediction problem for social networks," \textit{Journal of the American society for information science and technology}, vol. 58, no. 7, pp. 1019-1031, 2007.
\bibitem{b29} X. Wang, P. Cui, J. Wang, J. Pei, W. Zhu and S. Yang, ``Community preserving network embedding," \textit{in AAAI}, 2017.
\bibitem{b30} X. Yu, X. Ren, Y. Sun, Q. Gu, B. Sturt, U. Khandelwal, B. Norick and J. Han, ``Personalized entity recommendation: A heterogeneous information network approach," \textit{in WSDM}, 2014, pp. 283-292.
\bibitem{b31} D. Wang, P. Cui and W. Zhu, ``Structural deep network embedding," \textit{in SIGKDD}, 2016, pp. 1225-1234.

\bibitem{DBLP:conf/mdm/Kieu0J18}
T.~Kieu, B.~Yang, and C.~S. Jensen.
\newblock Outlier detection for multidimensional time series using deep neural
networks.
\newblock In {\em {MDM}}, pages 125--134, 2018.

\bibitem{DBLP:journals/vldb/YangDGJH18}
B.~Yang, J.~Dai, C.~Guo, C.~S. Jensen, and J.~Hu.
\newblock {PACE:} a path-centric paradigm for stochastic path finding.
\newblock {\em {VLDB} J.}, 27(2):153--178, 2018.

\bibitem{DBLP:journals/pvldb/0002GJ13}
B.~Yang, C.~Guo, and C.~S. Jensen.
\newblock Travel cost inference from sparse, spatio-temporally correlated time
series using markov models.
\newblock {\em {PVLDB}}, 6(9):769--780, 2013.

\bibitem{b32} L. van der Maaten and G. Hinton, ``Visualizing data using t-SNE," \textit{Journal of Machine Learning Research}, vol. 9, pp. 2579-2605, 2008.
\bibitem{b33} D. Erhan, Y. Bengio, A. Courville, P. Manzagol, P. Vincent and S. Bengio, ``Why does unsupervised pre-training help deep learning?" \textit{Jour. Machine Learning Research}, vol. 11, pp. 625-660, 2010.
\bibitem{DBLP:conf/cikm/CirsteaMMG018}
R.~Cirstea, D.~Micu, G.~Muresan, C.~Guo, and B.~Yang.
\newblock Correlated time series forecasting using multi-task deep neural
  networks.
\newblock In {\em {CIKM}}, pages 1527--1530, 2018.


\bibitem{DBLP:journals/tkde/YangKJ14}
B.~Yang, M.~Kaul, and C.~S. Jensen.
\newblock Using incomplete information for complete weight annotation of road
  networks.
\newblock {\em {IEEE} Trans. Knowl. Data Eng.}, 26(5):1267--1279, 2014.

\bibitem{DBLP:conf/icde/HuG0J19}
J.~Hu, C.~Guo, B.~Yang, and C.~S. Jensen.
\newblock Stochastic weight completion for road networks using graph
  convolutional networks.
\newblock In {\em {ICDE}}, pages 1274--1285, 2019.

\end{thebibliography}
\end{document}